\documentclass[11pt,a4paper]{article}
\pdfoutput=1
\usepackage{pslatex}
\usepackage{times,latexsym}
\usepackage{url}
\usepackage[T1]{fontenc}
\usepackage[acceptedWithA]{tacl2018v2}

\usepackage{graphicx}
\usepackage{amsfonts}
\usepackage{amsmath}
\usepackage{bbm}
\usepackage{array}
\usepackage{booktabs}
\usepackage{paralist}
\usepackage{multirow,multicol}
\usepackage{xcolor}

\graphicspath{{./figures/}}

\definecolor{darkgray2}{rgb}{0.36, 0.36, 0.36}
\newcommand{\U}[1]{\underline{#1}}

\newcommand{\light}[1]{\textcolor{darkgray2}{#1}}
\newcommand{\corpus}{\mbox{\sc Space}}
\newcommand{\yelp}{\mbox{\sc Yelp}}
\newcommand{\amazon}{\mbox{\sc Amazon}}
\newcommand{\subtext}[1]{\mbox{\footnotesize{$_{\mathrm{#1}}$}}}
\newcommand{\ssubtext}[1]{\mbox{\scriptsize{$_{\mathrm{#1}}$}}}

\DeclareMathOperator*{\argmin}{arg\,min}

\usepackage{microtype}

\title{Extractive Opinion Summarization in Quantized Transformer Spaces}

\author{
Stefanos Angelidis$^{1}$ \quad
Reinald Kim Amplayo$^{1}$ \\
\textbf{Yoshihiko Suhara$^{2}$ \quad
Xiaolan Wang$^{2}$ \quad
Mirella Lapata$^{1}$} \\
$^1$University of Edinburgh \quad
$^2$Megagon Labs \\
\texttt{s.angelidis@ed.ac.uk},~~\texttt{reinald.kim@ed.ac.uk}\\
\texttt{yoshi@megagon.ai},~~\texttt{xiaolan@megagon.ai}\\
\texttt{mlap@inf.ed.ac.uk}}

\date{}

\begin{document}

\maketitle

\begin{abstract}
  We present the Quantized Transformer (QT), an unsupervised system for
  extractive opinion summarization. QT is inspired by Vector-Quantized
  Variational Autoencoders, which we repurpose for popularity-driven
  summarization. It uses a clustering interpretation of the quantized space and
  a novel extraction algorithm to discover popular opinions among hundreds of
  reviews, a significant step towards opinion summarization of practical scope.
  In addition, QT enables controllable summarization without further training,
  by utilizing properties of the quantized space to extract aspect-specific
  summaries.  We also make publicly available \corpus, a large-scale evaluation
  benchmark for opinion summarizers, comprising general and aspect-specific
  summaries for 50~hotels.  Experiments demonstrate the promise of our approach, which
  is validated by human studies where judges showed clear preference for our
  method over competitive baselines.
\end{abstract}

\section{Introduction}
\label{sec:intro}

Online reviews play an integral role in modern life, as we look to previous
customer experiences to inform everyday decisions.  The need to digest review
content has fueled progress in opinion mining \cite{pang-lee-2008-opinion},
whose central goal is to automatically summarize people's attitudes towards an
entity. Early work \citep{hu-liu-2004-mining} focused on numerically aggregating
customer satisfaction across different \emph{aspects} of the entity under
consideration (e.g.,~the quality of a camera, its size, clarity).  More
recently, the success of neural summarizers in the Wikipedia and news domains
\citep{cheng-lapata-2016-neural,see-etal-2017-get,
narayan-etal-2018-ranking,liu2018generating,perez2019generating} has spurred
interest in \emph{opinion summarization}; the aggregation, in textual form, of
opinions expressed in a set of reviews
\citep{angelidis-lapata-2018-summarizing,huy-tien-etal-2019-opinions,
tian-etal-2019-aspect,coavoux-etal-2019-unsupervised,chu-liu-2019-meansum,
isonuma-etal-2019-unsupervised,
bravzinskas-etal-2020-unsupervised,amplayo-lapata-2020-unsupervised,
suhara-etal-2020-opiniondigest,wang2020extreme}.

Opinion summarization has distinct characteristics that set it apart
from other summarization tasks. Firstly, it cannot rely on reference
summaries for training, because such \emph{meta-reviews} are very
scarce and their crowdsourcing is unfeasible. Even for a single
entity, annotators would have to produce summaries after reading
hundreds, sometimes thousands, of reviews.  Secondly, the inherent
subjectivity of review text distorts the notion of information
\emph{importance} used in generic summarization
\cite{peyrard-2019-simple}.  Conflicting opinions are often expressed
for the same entity and, therefore, useful summaries should be based
on \emph{opinion popularity} \citep{ganesan-etal-2010-opinosis}.
Moreover, methods need to be \emph{flexible} with respect to the
size of the input (entities are frequently reviewed by thousands of
users), and \emph{controllable} with respect to the scope of the
output. For instance, users may wish to read a \textsl{general}
overview summary, or a more targeted one about a particular
\emph{aspect} of interest (e.g.,~a hotel's \textsl{location}, its
\textsl{cleanliness}, or available \textsl{food} options).

Recent work \citep{tian-etal-2019-aspect,
coavoux-etal-2019-unsupervised,chu-liu-2019-meansum,
isonuma-etal-2019-unsupervised,bravzinskas-etal-2020-unsupervised,
amplayo-lapata-2020-unsupervised,suhara-etal-2020-opiniondigest} has
increasingly focused on \emph{abstractive} summarization, where a summary is
generated token-by-token to create novel sentences that articulate
prevalent opinions in the input reviews. The abstractive approach
offers a solution to the lack of supervision, under the assumption that
opinion summaries should be written in the style of reviews. This
simplification has allowed abstractive models to generate
\emph{review-like} summaries from aggregate input representations,
using sequence-to-sequence models trained to reconstruct single
reviews. Despite being fluent, abstractive summaries may still suffer from
issues of text degeneration \citep{holtzman-etal-2019-curious},
hallucinations \citep{rohrbach-etal-2018-object}, and the undesirable
use of first-person narrative, a direct consequence of review-like
generation.  In addition, previous work used an unrealistically
small number of input reviews (10 or fewer), and only sparingly
investigated controllable summarization, albeit in weakly supervised
settings
\citep{amplayo-lapata-2019-informative,suhara-etal-2020-opiniondigest}.

In this paper, we attempt to address shortcomings of existing methods
by turning to \emph{extractive} summarization which aims to construct
an opinion summary by selecting a few representative input sentences
\citep{angelidis-lapata-2018-summarizing,huy-tien-etal-2019-opinions}.
Specifically, we introduce the \emph{Quantized Transformer} (QT), an
unsupervised neural model inspired by Vector-Quantized Variational
Autoencoders (\mbox{VQ-VAE};
\citealp{oord-etal-2017-neural,roy-etal-2018-theory}), which we
repurpose for popularity-driven summarization. QT combines
Transformers \citep{vaswani-etal-2017-attention} with the
discretization bottleneck of VQ-VAEs and is trained via sentence
reconstruction, similarly to the work of
\citet{roy-grangier-2019-unsupervised} on paraphrasing. At inference
time, we use a clustering interpretation of the quantized space and a
novel extraction algorithm that discovers popular opinions among
\emph{hundreds of reviews}, a significant step towards opinion
summarization of practical scope. QT is also capable of
aspect-specific summarization without further training, by exploiting
the properties of the Transformer's multi-head sentence
representations.

We further contribute to the progress of opinion mining research, by introducing
\corpus{} (shorthand for \U{S}ummaries of \U{P}opular and \U{A}spect-specific
\U{C}ustomer \U{E}xperiences), a large-scale corpus for the evaluation of
opinion summarizers. We collected 1,050 human-written summaries of
\mbox{TripAdvisor} reviews for 50 hotels. \corpus{} has \emph{general}
summaries, giving a high-level overview of popular opinions, and
\emph{aspect-specific} ones, providing detail on individual aspects (e.g.,
location, cleanliness). Each summary is based on 100 customer reviews, an order
of magnitude increase over existing corpora, thus providing a more realistic
input to competing models. Experiments on \corpus{} and two more benchmarks
demonstrate that our approach holds promise for opinion summarization.
Participants in human evaluation further express a clear preference for our
model over competitive baselines. We make \corpus{} and our code publicly
available.\footnote{\url{https://github.com/stangelid/qt}}

\section{Related Work}
\label{sec:related}
\citet{ganesan-etal-2010-opinosis} were the first to make the connection
between opinion mining and text summarization; they develop Opinosis,
a graph-based abstractive summarizer which explicitly models
\emph{opinion popularity}, a key characteristic of subjective text,
and central to our approach. Follow-on work
\cite{difabbrizio2014hybrid} adopts a hybrid approach where salient
sentences are first extracted and abstracts are generated based on
hand-written templates \cite{carenini2006multi}.  More recently,
\citet{angelidis-lapata-2018-summarizing} extract \emph{salient}
opinions according to their polarity intensity and aspect specificity,
in a weakly supervised setting.

A popular approach to modeling opinion popularity, albeit indirectly, is vector
averaging. \citet{chu-liu-2019-meansum} propose MeanSum, an unsupervised
abstractive summarizer that learns a review decoder through reconstruction, and
uses it to generate summaries conditioned on averaged representations of the
inputs. Averaging is also used by \citet{bravzinskas-etal-2020-unsupervised},
who train a \mbox{copy-enabled} variational autoencoder by reconstructing
reviews from averaged vectors of reviews about the same entity. Other methods
include denoising autoencoders \citep{amplayo-lapata-2020-unsupervised} and the
system of \citet{coavoux-etal-2019-unsupervised}, an encoder-decoder
architecture that uses a clustering of the encoding space to identify opinion
groups, similar to our work.

Our model builds on the Vector Quantized Variational Autoencoder
(VQ-VAE; \citealt{oord-etal-2017-neural}), a recently proposed
training technique for learning discrete latent variables, which aims
to overcome problems of posterior collapse and large variance
associated with Variational Autoencoders
\cite{Kingma2014AutoEncodingVB}. Like other related discretization
techniques \cite{Maddison2017TheCD,Kaiser2018DiscreteAF}, VQ-VAE
passes the encoder output through a \emph{discretization bottleneck}
using a neighbor lookup in the space of latent code embeddings. The
application of VQ-VAEs to opinion summarization is novel, to our
knowledge, as well as the proposed sentence extraction algorithm. Our
model does not depend on vector averaging, nor does it suffer from
information loss and hallucination. Furthermore, it can easily
accommodate a large number of input reviews. Within NLP, VQ-VAEs have
been previously applied to neural machine translation
\cite{roy-etal-2018-theory} and paraphrase generation
\cite{roy-grangier-2019-unsupervised}. Our work is closest to
\citet{roy-grangier-2019-unsupervised} in its use of a quantized
Transformer, however we adopt a different training algorithm (Soft EM;
\citealp{roy-etal-2018-theory}), orders of magnitude fewer discrete
latent codes, a different method for obtaining head sentence vectors,
and apply the QT in a novel way for extractive opinion summarization.

Besides modeling, our work contributes to the growing body of resources for
opinion summarization. We release \corpus{}, the first corpus to contain both
general and aspect-specific opinion summaries, while increasing the number of
input reviews tenfold compared to popular benchmarks
\cite{bravzinskas-etal-2020-unsupervised,chu-liu-2019-meansum,
angelidis-lapata-2018-summarizing}.

\section{Problem Formulation}
\label{sec:method}

Let~$C$ be a corpus of reviews on entities $\{e_1, e_2, \dots\}$ from a single
domain~$d$, e.g.,~{hotels}. Reviews may discuss any number of relevant
aspects \mbox{$A_d = \{a_1, a_2, \dots\}$}, like the hotel's \textsl{rooms} or
\textsl{location}. For every entity~$e$, we define its review set \mbox{$R_e =
\{r_1, r_2, \dots\}$}. Every review is a sequence of sentences \mbox{$(x_1, x_2,
\dots)$} and a sentence $x$ is, in turn, a sequence of words \mbox{$(w_1, w_2,
\dots)$}.  For brevity, we use $X_e$ to denote all review sentences about
entity~$e$. We formalize two sub-tasks: (a)~\emph{general opinion
summarization}, where a summary should cover popular opinions in $R_e$ across
all discussed aspects; and (b) \emph{aspect opinion summarization}, where a
summary must focus on a single specified aspect $a \in A_d$. In our extractive
setting, these translate to creating a general or aspect summary by selecting a
small subset of sentences $S_e\subset X_e$.
 
We train the Quantized Transformer (QT) through sentence reconstruction to learn
a rich representation space and its quantization into latent codes
(Section~\ref{sec:method-qt}). We enable opinion summarization, by mapping input
sentences onto their nearest latent codes and extract those sentences that are
representative of the most popular codes (Section~\ref{sec:method-summ}). We
also illustrate how to produce aspect-specific summaries using a trained QT
model and a few aspect-denoting query terms
(Section~\ref{sec:aspect-opin-summ}).

\subsection{The Quantized Transformer}
\label{sec:method-qt}

Our model is a variant of VQ-VAEs
\citep{oord-etal-2017-neural,roy-grangier-2019-unsupervised} and
consists of: (a) a Transformer sentence \emph{encoder} which encodes
an input sentence $x$ into a multi-head representation
$\{\mathrm{x}_1, \dots, \mathrm{x}_H\}$, where
$\mathrm{x}_h \in \mathbb{R}^D$ and $H$ is the number of heads; (b) a vector
\emph{quantizer} that maps each head vector to a mixture of discrete latent
codes, and uses the codes' embeddings to produce quantized vectors
$\{\mathrm{q}_1, \dots,\mathrm{q}_H\}$, $\mathrm{q}_h \in \mathbb{R}^D$; (c) a
Transformer sentence \emph{decoder}, which attends over the quantized vectors
to generate sentence reconstruction $\hat{x}$. The decoder is not used during
summarization; we only use the learned quantized space to extract
sentences, as described in Section~\ref{sec:method-summ}.

\begin{figure}[t]
  \centering
  \includegraphics[width=\columnwidth]{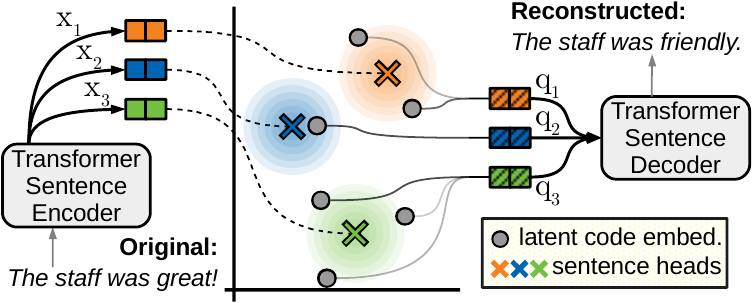}
  \caption{A sentence is encoded into a 3-head representation and head
    vectors are quantized using a weighted average of their
    neighboring code embeddings. The QT model is trained by
    reconstructing the original sentence.}
  \label{fig:qt-train}
\end{figure}

\paragraph{Sentence Encoding}
Our encoder prepends sentence $x$ with the special token
\texttt{[SNT]} and uses the vanilla Transformer encoder
\cite{vaswani-etal-2017-attention} to produce token-level vectors. We
ignore individual word vectors and only keep the special token's
vector $\mathrm{x}_{snt} \in \mathbb{R}^D$. We obtain a multi-head
representation of $x$, by splitting $\mathrm{x}_{snt}$ into $H$
sub-vectors \mbox{$\{\mathrm{x}'_1, \dots, \mathrm{x}'_H\}$,
  $\mathrm{x}'_h \in \mathbb{R}^{D/H}$}, followed by a
\mbox{layer-normalized} transformation:
\begin{equation}
  \mathrm{x}_h = \mathrm{LayerNorm}(\mathrm{Wx}'_h + \mathrm{b}) \,,
\end{equation}
where $\mathrm{x}_h$ is the $h$-th head and \mbox{$\mathrm{W} \in
  \mathbb{R}^{D \times D/H}$}, \mbox{$b \in \mathbb{R}^D$} are shared
across heads.  Hyperparameter $H$, i.e.,~the number of \emph{sentence
  heads} of our encoder, is different from Transformer's internal
\emph{attention heads}.  The encoder's operation is illustrated in
Figure \ref{fig:qt-train}, where the sentence ``\textsl{The staff was
  great!}" is encoded into a 3-head representation.

\paragraph{Vector Quantization}
Let $z_1, \dots, z_H$ be discrete latent variables corresponding to $H$~encoder
heads. Every variable can take one of $K$ possible latent codes, $z_h \in [K]$.
The quantizer's \emph{codebook}, \mbox{$\mathrm{e} \in \mathbb{R}^{K \times D}$}, is
shared across latent variables and maps each code (or \emph{cluster}) to
its embedding (or \emph{centroid}) $\mathrm{e}_k \in \mathbb{R}^D$. Given
sentence~$x$ and its multi-head encoding $\{\mathrm{x}_1, \dots,
\mathrm{x}_H\}$, we independently quantize every head using a mixture of its
nearest codes from $[K]$. Specifically, we follow the \emph{Soft EM}
training of \citet{roy-etal-2018-theory} and sample, with replacement,
$m$ latent codes for the $h$-th head:%
\begin{equation} 
  \begin{split}
    z_h^1, \dots, z_h^m &\sim \mathrm{Multinomial}(l_1, \dots, l_K) \,,\\ 
    \mathrm{with} &\; l_k = -\|\mathrm{x}_h - \mathrm{e}_k\|_2^2 \,, 
    \label{eq:sample}
  \end{split}
\end{equation}
where $\mathrm{Multinomial}(l_1, \dots, l_K)$ is a K-way multinomial
distribution with logits $l_1, \dots, l_K$. The $h$-th quantized head vector is
obtained as the average of the sampled codes' embeddings: \begin{equation}
\mathrm{q}_h = \frac{1}{m} \sum_{j=1}^m \mathrm{e}_{z_h^j} \,.  \end{equation}
This soft quantization process is shown in Figure~\ref{fig:qt-train}, where
head vectors $\mathrm{x}_1$, $\mathrm{x}_2$ and $\mathrm{x}_3$ are quantized
using a weighted average of their neighboring code embeddings, to produce
$\mathrm{q}_1$, $\mathrm{q}_2$, and $\mathrm{q}_3$.

\paragraph{Sentence Reconstruction and Training}
Instead of attending over individual token vectors, as in the vanilla
architecture, the Transformer sentence decoder attends over 
$\{\mathrm{q}_1, \dots,\mathrm{q}_H\}$, the quantized head
vectors of the sentence, to generate reconstruction~$\hat{x}$. The model is
trained to minimize:
\begin{equation}
  L = L_r + \sum_{h}\|\mathrm{x}_{h} - \mathrm{sg}(\mathrm{q}_{h})\|_2 \,.
\end{equation}
\noindent $L_r$ is the reconstruction cross entropy, and stop-gradient
operator $\mathrm{sg}(\cdot)$ is defined as identity during forward
computation and zero on backpropagation. The sampling of
Equation~(\ref{eq:sample}) is bypassed using the straight-through
estimator \citep{bengio-etal-2013-estimating} and the latent codebook
is trained via exponentially moving averages, as detailed in
\citet{roy-etal-2018-theory}.

\subsection{Summarization in Quantized Spaces}
\label{sec:method-summ}

Existing neural methods for opinion summarization have modeled opinion
popularity within a set of reviews by encoding each review into a vector,
averaging all vectors to obtain an aggregate representation of the input, and
feeding it to a review decoder to produce a summary
\citep{chu-liu-2019-meansum,coavoux-etal-2019-unsupervised,
bravzinskas-etal-2020-unsupervised}. This approach is problematic for two
reasons. Firstly, it assumes that complex semantics of whole reviews can be
encoded in a single vector. Secondly, it also assumes that features of commonly
occurring opinions are preserved after averaging and, therefore, those opinions
will appear in the generated summary. The latter assumption becomes particularly
uncertain for larger numbers of input reviews.

We take a different approach, using sentences as the unit of
representation, and propose a general extraction algorithm based on
the QT, which explicitly models popularity without vector
aggregation. Using the same algorithmic framework we are also able to
extract aspect-specific summaries.

\subsubsection{General Opinion Summarization}
\label{sec:gener-opin-summ}

We exploit QT's quantization of the encoding space to cluster similar sentences
together, quantify the popularity of the resulting clusters, and extract representative
sentences from the most popular ones.

\begin{figure}[t]
  \centering
  \includegraphics[width=\columnwidth]{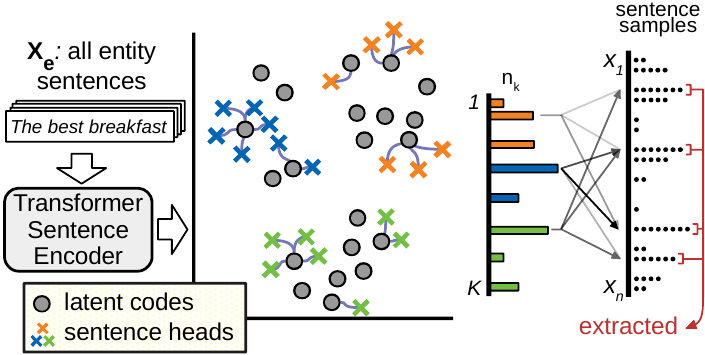}
  \caption{General opinion summarization with QT. All input sentences for an
    entity are encoded using three heads (shown in orange, blue and green
    crosses). Sentence vectors are clustered under their nearest latent code
    (gray circles). Popular clusters (histogram) correspond to commonly occurring
  opinions, and are used to sample and extract the most representative sentences.}
  \label{fig:qt-summ}
\end{figure}

Specifically, given $X_e = \{x_1, \dots, x_i, \dots, x_N\}$, the $N$ review
sentences about entity $e$, the trained encoder produces $N \times H$
unquantized head vectors \mbox{$\{\mathrm{x}_{11}, \dots, \mathrm{x}_{ih},
\dots, \mathrm{x}_{NH}\}$}, where $\mathrm{x}_{ih}$ is the $h$-th head of the
$i$-th sentence. We perform \emph{hard} quantization, assigning every vector
to its nearest latent code, and counting the number of assignments per code,
i.e., the \emph{popularity} of each cluster:
\begin{align}
  z_{ih} &= \argmin_{k \in [K]} \|\mathrm{x}_{ih} - \mathrm{e}_k\|_2 \label{eq:assign} \\
  n_k &= \sum_{i,h}\mathbbm{1}[z_{ih} = k] \,. \label{eq:count}
\end{align}
Figure~\ref{fig:qt-summ} shows how sentences $X_e$ are encoded, and
their different heads are assigned to codes. Similar sentences cluster under
the same codes and, consequently, clusters receiving numerous assignments are
characteristic of popular opinions in $X_e$. A general summary should consist of
the sentences that are most \emph{representative} of these popular clusters.

In the simplest case, we could couple every code~$k$ with its nearest
sentence~$x^{(k)}$:
\begin{equation}
  x^{(k)} = \argmin_{i}(\min_{h}\|\mathrm{x}_{ih} - \mathrm{e}_k\|_2) \,, 
  \label{eq:couple}
\end{equation}
\noindent and rank sentences $x^{(k)}$ according to the
size~$n_k$ of their respective clusters; the top sentences, up to a predefined
budget, are extracted into a summary.  

\begin{figure}[t]
  \centering
  \includegraphics[width=\columnwidth]{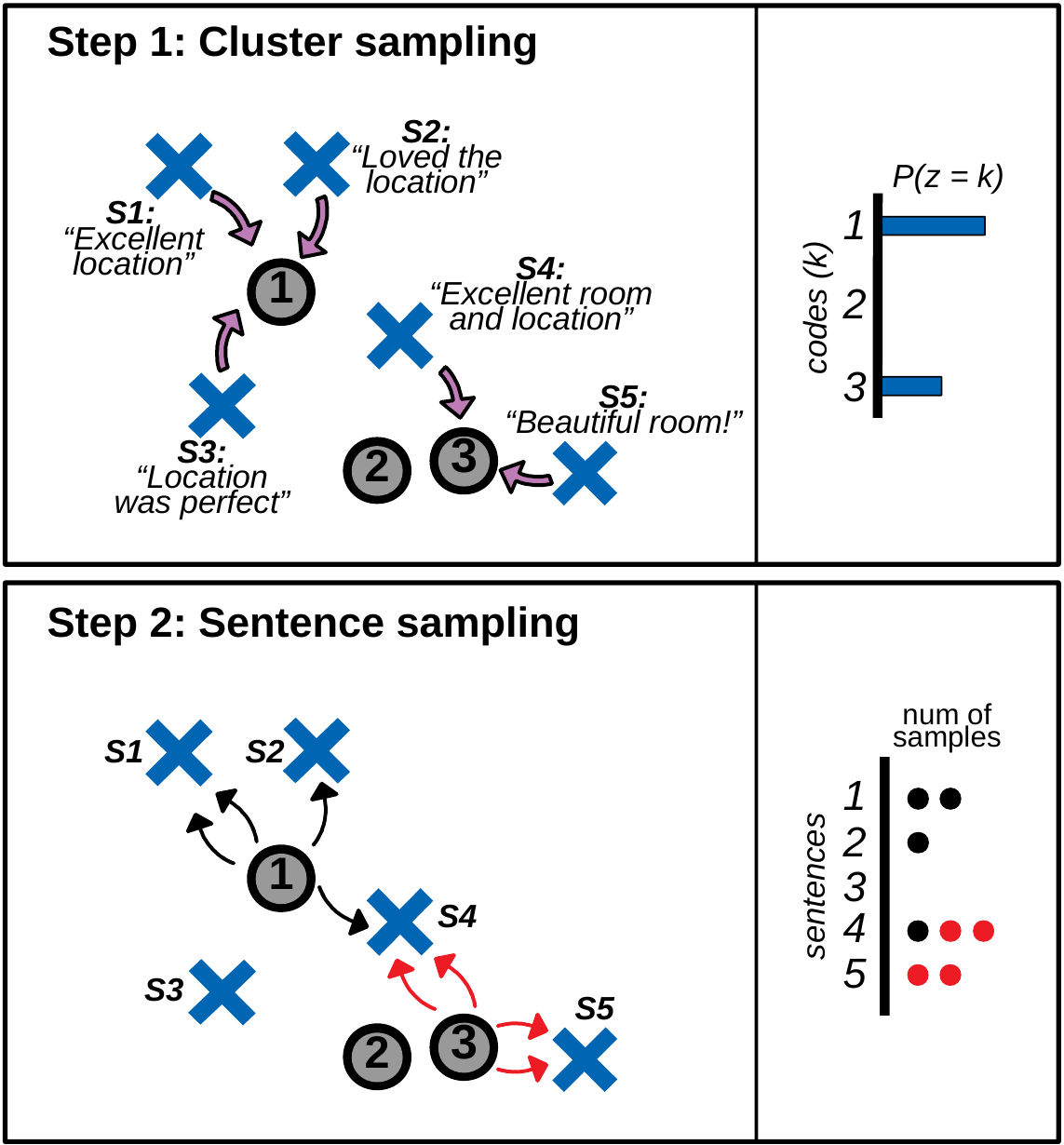}
  \begin{picture}(0,0)
    \put(60,241){\scriptsize{$\mathsf{Probability}$}}
    \put(49,233){\scriptsize{$\mathsf{of~sampling~code}$}}
    \put(96,200.5){\tiny{$\mathbf{3/5}$}}
    \put(87,168){\tiny{$\mathbf{2/5}$}}
    \put(54,122){\scriptsize{$\mathsf{Samples~drawn}$}}
    \put(50,114){\scriptsize{$\mathsf{after~2~iterations}$}}
  \end{picture}
  \vspace{-5mm}
  \caption{Sentence ranking via two-step sampling. In this toy example, each
    sentence ($s_1$ to $s_5$) is assigned to its nearest code (\mbox{$k = 1,\,
    2,\, 3$}), as shown by thick purple arrows. During \emph{cluster sampling},
    the probability of sampling a code (top right; shown as blue bars) is
    proportional to the number of assignments it receives. For every sampled
    code, we perform \emph{sentence sampling}; sentences are sampled, with
    replacement, according to their proximity to the code's encoding. Samples
    from codes 1 and 3 are shown in black and red, respectively.}
  \label{fig:sample}
\end{figure}
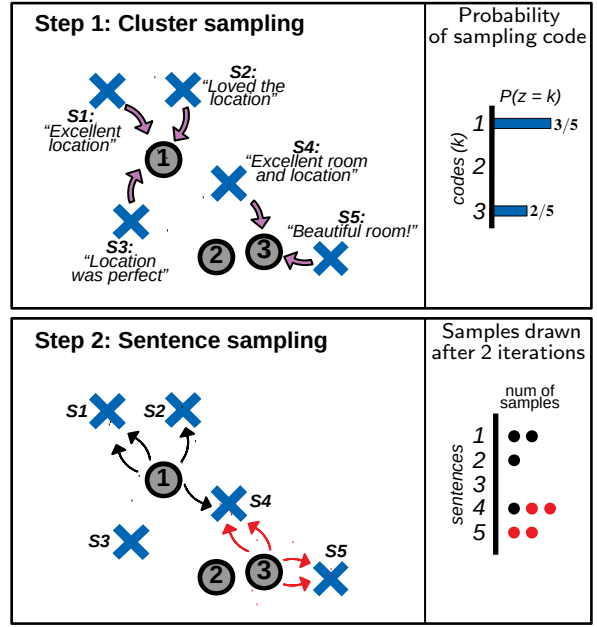

The above ranking method entails that only those sentences which are the nearest
neighbor of a popular code are likely to be extracted. However, a
salient sentence may be in the neighborhood of multiple codes per
head, despite never being the \emph{nearest} sentence of a code vector. For
example, the sentence \textsl{``Great location and beautiful rooms''}
is representative of clusters encoding positive attitudes for both the
\textsl{location} and the \textsl{rooms} of a hotel. To capture this,
we relax the requirement of coupling every cluster with exactly one
sentence and propose \emph{two-step sampling} (Figure~\ref{fig:sample}), a novel
sampling process which simultaneously estimates cluster popularity and promotes
sentences commonly found in the proximity of popular clusters.  We repeatedly
perform the following operations:

\paragraph{Cluster Sampling} We first sample a latent code~$z$ with
probability proportional to the clusters' size:%
\begin{equation}
  z \sim P(z=k) = \frac{n_k}{N \times H} \,,
\end{equation}  
\noindent where $n_k$ is the number of assignments for code $k$, 
computed in Equation~(\ref{eq:count}). For example, if the input
contains many paraphrases of sentence \textsl{``Excellent
location''}, these are likely to be clustered under the same code,
which in turn increases the probability of sampling that code. Cluster
sampling is illustrated on the top of Figure~\ref{fig:sample}, showing
assignments (left) and resulting code probabilities (right).

\paragraph{Sentence Sampling} The sampled code $z$ exists in the
neighborhood of many input sentences. Picking a single sentence as
the most characteristic of that cluster is too restrictive. Instead,
we sample (with replacement) sentences from the code's neighborhood
$n$ times, thus generalizing Equation~(\ref{eq:couple}):%
\begin{equation}
  \begin{split}
    x^1, \dots, x^n &\sim \mathrm{Multinomial}(l'_1, \dots, l'_N) \,, \\
    \mathrm{with} &\; l'_i = -\min_h(\|\mathrm{x}_{ih} - \mathrm{e}_z\|_2^2) \,,
  \end{split}
  \label{eq:sent-sample}
\end{equation}
\noindent where the Multinomial's logits~$l'_i$ mark the (negative)
distance of the $i$-th sentence's head which is closest to
code~$z$. Sentence sampling is depicted in the toy example of
Figure~\ref{fig:sample} (bottom). After selecting code $k = 1$ during
cluster sampling, four sentence samples are drawn (shown in black
arrows). The next cluster sample ($k = 3$) results in four more
sentence samples (shown in red). Sentence $s_4$ (\textsl{``Excellent
  room and location''}) receives the most \emph{votes} in total, after
being sampled as a neighbor of both codes.

Two-step sampling is repeated multiple times and all sentences in~$X_e$ are
ranked according to the total number of times they have been sampled. The final
summary is constructed by concatenating the top ones (see right part of 
Figure~\ref{fig:qt-summ}). Importantly, our extraction algorithm is not
sensitive to the size of the input. More sentences increase the absolute number
of assignments per code, but do not hinder two-step sampling or cause
information loss; on the contrary, a larger pool of sentences may result in a
more densely populated quantized encoding space and, in turn, a
better estimation of cluster popularity and sentence ranking.

\subsubsection{Aspect Opinion Summarization}
\label{sec:aspect-opin-summ}

So far, we have focused on selecting sentences solely based on the
popularity of the opinions they express.  We now turn our attention to
aspect summaries, which discuss a particular aspect of an entity
(e.g., the \textsl{location} or \textsl{service} of a hotel) while
still presenting popular opinions. We create such summaries with a
trained QT model, without additional fine-tuning. Instead, we exploit
QT's multi-head representations and only require a small number of
aspect-denoting query terms.\footnote{Contrary to
  \citet{angelidis-lapata-2018-summarizing} who used 30 seed-words per
  aspect, we only assume five query terms per aspect to replicate a
  realistic setting.}

\begin{figure}[t]
  \centering
  \includegraphics[width=\columnwidth]{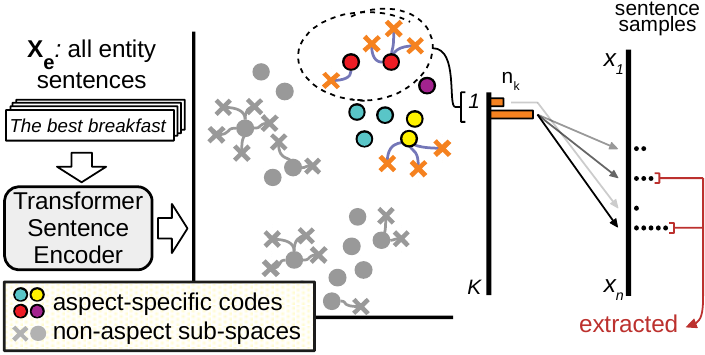}
  \caption{Aspect opinion summarization with QT. The aspect-encoding
    sub-space is identified using mean aspect entropy and all other
    sub-spaces are ignored (shown in gray). Two-step sampling is restricted only
    to the codes associated with the desired aspect (shown in red).}
  \label{fig:qt-asp-summ}
\end{figure}

We hypothesize that different sentence heads in QT encode the
approximately orthogonal semantic or structural attributes which are
necessary for sentence reconstruction. In the simplified example in
Figure~\ref{fig:qt-summ}, the encoder's first head (orange) might
capture information about the aspects of the sentence, the second head
(blue) encodes sentiment, while head three (green) may encode
structural information (e.g., the length of the sentence or its
punctuation). Our hypothesis is reinforced by the empirical
observation that sentence vectors originating from the same head will
occupy their own \emph{sub-space}, and do not show any similarity to
vectors from other heads. As a result, each latent code $k$ receives
assignments from exactly one head of the sentence encoder. More
formally, head $h$ yields a set of latent codes such that \mbox{$K_h
  \subset [K]$}. Figure~\ref{fig:qt-summ} demonstrates this, as the
encoding space consists of three sub-spaces, one for each head.
Sentence and latent code vectors are further organized within that
sub-space according to the attribute captured by the respective head.

\begin{table*}[t]
  \centering
  \footnotesize
  \begin{tabular}{@{}l|r|rrr|cc@{}}
    \toprule[1pt]
    & \multicolumn{1}{c|}{Reviews} & \multicolumn{1}{c}{Entities} & \multicolumn{1}{c}{Rev/Ent}  & \multicolumn{1}{c}{Summaries} (R) & \multicolumn{1}{|c}{Type} & \multicolumn{1}{c}{Scope} \\
    \midrule[1pt]
    \corpus{} (This work)
      & 1.14M & 50 & 100 & 1,050 (3) & Abstractive & General+Aspect \\

    \amazon{} \citep{bravzinskas-etal-2020-unsupervised}
      & 4.75M & 60 & 8 & 180 (3) & Abstractive & General only \\

    \yelp{} \citep{chu-liu-2019-meansum}
      & 1.29M & 200 & 8 & 200 (1) & Abstractive & General only \\

    \textsc{OpoSum} \citep{angelidis-lapata-2018-summarizing} $\!\!\!\!$
      & 359K & 60 & 10 & 180 (3) & Extractive & General only \\

    \bottomrule[1pt]
  \end{tabular}
  \caption{Statistics for \corpus{} and three recently introduced
    evaluation corpora for opinion summarization. \corpus{} includes
    aspect summaries for six aspects.  (Reviews: number of reviews in
    \emph{training} set, no gold-standard summaries are available;
    Rev/Ent: Input reviews per entity in \emph{test set}; R:
    Reference summaries per example).}
  \label{tab:datastats}
\end{table*}

To enable aspect summarization, we identify the sub-space capturing
aspect-relevant information and label its aspect-specific codes, as
seen in Figure~\ref{fig:qt-asp-summ}.  Specifically, we first quantify
the probability of finding an aspect in the sentences assigned to a
latent code and identify the head sub-space that best separates
sentences according to their aspect. Then, we map every cluster within
that sub-space to an aspect and extract aspect summaries only from those
aspect-specific clusters.

We utilize a held-out set of review sentences $X_{\textit{dev}}$, and
keywords \mbox{$Q_a = \{s_1, \dots, s_5\}$} for aspect~$a$. We encode
and quantize sentences in $X_{\textit{dev}}$ and compute the
probability that latent code $k$ contains tokens typical of aspect $a$ as:
\begin{equation}
  P_k(a) = \frac{\mathit{tf}(Q_a,k)}{\sum\limits_{a'}\mathit{tf}(Q_{a'},k)} \,,
  \label{eq:asp-prob}\vspace{-1mm}
\end{equation}
\noindent where $\mathit{tf}(Q_a,k)$ is the number of times query
terms in $Q_a$ where found in sentences assigned to~$k$. We use
information theory's \emph{entropy} to measure how aspect-certain
code~$k$ is:
\begin{equation}
  \mathcal{H}_k = -\sum_aP_k(a)\,\mathrm{log}P_k(a) \,.
\end{equation}
\noindent Low aspect entropy values indicate that most sentences
assigned to~$k$ belong to a single aspect. It thus follows that
$h^{\mathit{asp}}$, i.e., the head sub-space which best separates
sentences according to their aspect, will exhibit the lowest
{mean aspect entropy}:
\begin{equation}
  h^{\mathit{asp}} = \argmin_h \left( \frac{1}{|K_h|} \sum_{k \in K_h} \mathcal{H}_k \right) \,.
\end{equation}
\noindent We map every code produced by $h^\mathit{asp}$ to its aspect
$a^{(k)}$ via Equation (\ref{eq:asp-prob}), and obtain aspect codes:
\begin{equation}
  K_a = 
  \{k \, | \, k \in K_{h^\mathit{asp}} \; \mathrm{and} \; a = a^{(k)} \} \,.
  \label{eq:asp-codes}
\end{equation}
\noindent To extract a summary for aspect $a$, we follow the ranking
or sampling methods described in
Equations~(\ref{eq:assign})--(\ref{eq:sent-sample}), restricting the
process to codes $K_a$. Sub-space selection and aspect-specific
sentence sampling are illustrated in Figure~\ref{fig:qt-asp-summ}.

\section{The \corpus{} Corpus}
\label{sec:corpus}

We introduce \corpus{} (\U{S}ummaries of \U{P}opular and
\U{A}spect-specific \U{C}ustomer \U{E}xperiences), a large-scale
opinion summarization benchmark for the evaluation of unsupervised summarizers. 
\corpus{} is built on TripAdvisor hotel reviews and aims
to facilitate future research by improving upon the shortcomings of
existing datasets. It comes with a training set of approximately 1.1
million reviews for over 11 thousand hotels, obtained by cleaning and
downsampling an existing collection \citep{wang-etal-2010-latent}.
The training set contains no reference summaries, and is useful for
unsupervised training.

For evaluation, we created a large collection of
human-written, abstractive opinion summaries. Specifically, for a
held-out set of 50 hotels (25 hotels for development and 25 for
testing), we asked human annotators to write high-level
{general} summaries \emph{and} {aspect} summaries for
six popular aspects: \textsl{building}, \textsl{cleanliness},
\textsl{food}, \textsl{location}, \textsl{rooms}, and
\textsl{service}. For every hotel and summary type, we collected three
reference summaries from different annotators.
Importantly, for every hotel, summaries were based on 100 input
reviews. To the best of our knowledge, this is the largest
crowdsourcing effort towards obtaining high-quality abstractive
summaries of reviews, and the first to use a pool of input reviews of
this scale (see Table~\ref{tab:datastats} for a comparison with
existing datasets).  Moreover, \corpus{} is the first benchmark to
also contain aspect-specific opinion summaries. 

The large number of input reviews per entity poses certain challenges
with regard to the collection of human summaries. A direct approach is
prohibitive, as it would require annotators to read all 100 reviews
and write a summary in a single step. A more reasonable method is to
first identify a subset of input sentences that most people consider
salient, and then ask annotators to summarize them. Summaries were
thus created in multiple stages using the
\mbox{Appen}\footnote{\url{https://appen.com/}} platform and expert annotator
channels of native English speakers. Although we propose an extractive
model, annotators were asked to produce abstractive summaries, as we
hope \corpus{} will be broadly useful to the summarization community. We did not
allow the use of first-person narrative to collect more summary-like texts. We
present our annotation procedure below.\footnote{Full annotation instructions:
\url{https://github.com/stangelid/qt/blob/main/annotation.md}}

\subsection{Sentence Selection via Voting}

The sentence selection stage identifies a subset of review sentences
which contain the most salient and useful opinions expressed by the
reviewers. This is a crucial but subjective task and, therefore, we
devised a voting scheme which allowed us to select sentences that
received votes by many annotators.

Specifically, each review was shown to five judges who were asked to select
informative sentences.  Annotators were encouraged to exercise their own
judgement in selecting summary-worthy sentences, but were advised to focus on
sentences which explicitly expressed or supported reviewer opinions, avoiding
overly general or personal comments (e.g., \textit{``Loved the hotel''},
\textit{``I like a shower with good pressure''}), and making sure that important
aspects were included. We set no threshold on the number of sentences they
could select (we allowed selecting \textsl{all} or \textsl{no} sentences for
a given review).  However, the annotation interface kept track of their total
votes and guided them to select between 20\% and 40\% of sentences, on average.

Sentences with~4 or more votes were automatically promoted to the next stage.
Inter-annotator agreement according to Cohen's kappa was $k=0.36$, indicating
{``fair agreement''}. Previous studies have shown that human agreement for
sentence selection tasks in summarization of news articles is usually lower than
0.3 \citep{radev-etal-2003-evaluation}. The median number of sentences promoted
for summarization for each hotel was 83, while the minimum was 46. This ensured
that enough sentences were always available for summarization, while
simplifying the task; annotators were now
required to read and summarize considerably smaller amounts of review text than
the original 100 reviews.

\subsection{Summary Collection}

\paragraph{General Summaries}
The top-voted sentences for each hotel were presented to three
annotators, who were asked to read them and produce a high-level
overview summary up to a budget of 100 words. To simplify the task and
help annotators write coherent summaries, sentences with high lexical
overlap were grouped together and the interface allowed the annotators
to quickly sort sentences according to words they contained. The
process resulted in an inter-annotator ROUGE-L score of~29.19
and provides ample room for future research, as detailed in our
experiments (Table~\ref{tab:space-general}).

\paragraph{Aspect Summaries}

Top-voted sentences were further labeled by an off-the-shelf aspect classifier
\citep{angelidis-lapata-2018-summarizing} trained on an public aspect-labeled
corpus of hotel review sentences \citep{marcheggiani2014hier}.\footnote{The
classifier's precision on the aspect-labeled corpus' development set is 85.4\%.} Sentences
outside of the six most popular aspects (\textsl{building},
\textsl{cleanliness}, \textsl{food}, \textsl{location}, \textsl{rooms}, and
\textsl{service}) were ignored, and sentences with 3 votes were promoted, only if an
aspect had no sentences with 4 votes. The promoted sentences were grouped
according to their aspect and presented to annotators, who were asked to
create a more detailed, aspect-specific summary, up to a budget of 75 words. The
aspect summaries have an inter-annotator \mbox{ROUGE-L} score of
34.58.

\section{Evaluation}
\label{sec:eval}

In this section, we discuss our experimental setup, including datasets
and comparison models, before presenting our automatic evaluation
results, human studies, and further analyses.

\subsection{Experimental Setup}

\paragraph{Datasets} 
We used \corpus{} as the main testbed for our experimental evaluation, covering
both general and aspect-specific summarization tasks. For general summarization,
we used two additional  opinion summarization benchmarks, namely
\yelp{} \citep{chu-liu-2019-meansum} and \amazon{}
\citep{bravzinskas-etal-2020-unsupervised} (see Table~\ref{tab:datastats}). For
all datasets, we use pre-defined development and test set splits, and only
report results on the test set.

\paragraph{Implementation Details}
We used unigram LM SentencePiece vocabularies of
32K.\footnote{\url{https://github.com/google/sentencepiece}} All system
hyperparameters were selected on the development set. The Transformer's
dimensionality was set to 320 and its feed-forward layer to 512.  We used 3
layers and 4 internal attention heads for its encoder and decoder, whose input
embedding layer was shared, but no positional encodings as we observed no
summarization improvements. We used $H=8$ sentence heads for representing every
sentence. For the quantizer, we set the number of latent codes to $K=1,024$ and
sampled $m=30$ codes for every input sentence, during training.  We used the
Adam optimizer, with initial learning rate of $10^{-3}$ and a learning rate
decay of~0.9. We warmed up the Transformer by disabling quantization for the
first 4 epochs. In total, we ran 20 training epochs. On the full \corpus{}
corpus, QT was trained in 4 days on a single GeForce GTX 1080 Ti GPU, using our
available PyTorch implementation.  All general and aspect summaries were
extracted with the two-step sampling procedure described in
Section~\ref{sec:gener-opin-summ}, unless otherwise stated.  When two-step
sampling was enabled, we ranked sentences by sampling 300 latent codes and, for
every code, sampled $n=30$ neighboring sentences.  QT and all extractive
baselines use a greedy algorithm to eliminate redundancy, similar to previous
research on multi-document summarization \citep{cao2015ranking,
yasunaga-etal-2017-graph,angelidis-lapata-2018-summarizing}.

\subsection{Metrics}
We evaluate the lexical overlap between system and human summaries
using ROUGE F-scores.\footnote{\url{https://github.com/bheinzerling/pyrouge}}
We report uni- and bi-gram variants (R1/R2), as well as longest common
subsequence (RL).

A successful opinion summarizer must also produce summaries which match
human-written ones in terms of aspects mentioned and sentiment
conveyed. For this reason, we also evaluate our systems
on two metrics which utilize an off-the-shelf aspect-based sentiment
analysis (ABSA) system \citep{miao2020snippext}, pre-trained in-domain.
The ABSA system extracts opinion phrases from summaries,
and predicts their aspect category and sentiment. The metrics use these
predictions as follows.
\begin{itemize}
  \itemsep0mm
\item[\textbf{Aspect Coverage}] We use the phrase-level aspect predictions to
  mark the presence or absence of an aspect in a summary. We discard very
  infrequent aspect categories. Similar to \citet{pan2020large}, we measure
  precision, recall, and F1 of system against human summaries. 
\item[\textbf{Aspect-level Sentiment}] We propose a new metric to
    evaluate the sentiment consistency between system and human
    summaries. Specifically, we compute the sentiment polarity score
    towards an individual aspect $a$ as the mean polarity of the
    opinion phrases that discuss this aspect in a summary (\mbox{$\mathit{pol}_a
      \in [-1, 1]$}). We repeat the process for every
    aspect, thus obtaining a vector of aspect polarities for the
    summary (we set the polarity of absent aspects to zero). The aspect-level
    sentiment consistency is computed as the mean squared error
    between system and human polarity vectors.
\end{itemize}

\subsection{Results: General Summarization}
We first discuss our results on general summarization and then move on
to present experiments on aspect-specific summarization. We compared
our model against the following baselines:

\begin{itemize}
  \itemsep0mm
\item[\textbf{Best Review}] systems select the single review that best
  approximates the consensus opinions in the input. We use a
  \textsl{Centroid} method that encodes the entity's reviews with BERT
  (average token vector; \citealp{devlin-etal-2019-bert}) or
  SentiNeuron \citep{radford-etal-2017-learning}, and picks the one
  closest to the mean review vector. We also tested an \textsl{Oracle}
  method, which selects the review closest to the reference summaries.

\item[\textbf{Extractive}] systems, where we tested \textsl{LexRank}
  \citep{erkan-radev-2004-lexrank}, an unsupervised graph-based
  summarizer. To compute its adjacency matrices, we used BERT and
  SentiNeuron vectors, in addition to the sparse tf-idf features of
  the original. We also present a \textsl{random} extractive baseline.

\item[\textbf{Abstractive}] systems include \textsl{Opinosis}
  \citep{ganesan-etal-2010-opinosis} a graph-based method;
  \textsl{MeanSum} \citep{chu-liu-2019-meansum}, and \textsl{Copycat}
  \citep{bravzinskas-etal-2020-unsupervised} two neural abstractive
  methods that generate review-like summaries from aggregate review
  representations learned using autoencoders.
\end{itemize}

\begin{table*}[t]
	\footnotesize
	\centering
\begin{tabular}{@{\hspace*{-.2cm}}ll@{}}
\begin{minipage}[b]{3.63in}
  \begin{tabular}{@{}cl@{}|@{}c@{~}r@{~~}c@{~}|@{~}c@{~}|@{~}c@{~}c@{~}c@{~}|@{~}c@{}}
    \toprule[1pt]
    \multicolumn{2}{l@{~}|}{\corpus\ \scriptsize{[\textsc{General}]}} &  R1 & \multicolumn{1}{c}{R2} 
      & RL & RL\ssubtext{ASP} & AC\ssubtext{P} & AC\ssubtext{R} & AC\ssubtext{F1} & SC\ssubtext{MSE} \\
    \midrule[1pt]

    \parbox[t]{1mm}{\multirow{4}{*}{\rotatebox[origin=c]{90}{\scriptsize{Best Review}}}}
    & Centroid\ssubtext{SENTI} & ~27.36 & 5.81  & 15.15 & \light{ 8.77} &.788&\textbf{.705}&\U{.744}&.580\\
    & Centroid\ssubtext{BERT}  & ~31.33 & 5.78  & 16.54 & \light{ 9.35} &.805&\U{.701}&\U{.749}&.524\\
    & Oracle\ssubtext{SENTI}   & ~\light{32.14} & \light{7.52} & \light{17.43}& \light{9.29} &\light{.817}&\light{.699}&\light{.753}&\light{.455}\\
    & Oracle\ssubtext{BERT}    & ~\light{33.21} & \light{8.33} & \light{18.02}& \light{9.67} &\light{.823}&\light{.777}&\light{.799}&\light{.401}\\
    \midrule                                                
                                                            
    \parbox[t]{1mm}{\multirow{4}{*}{\rotatebox[origin=c]{90}{\scriptsize{Extract}}}}
    & Random                  & ~26.24 & 3.58  & 14.72 & \light{11.53} &.799&.374&.509&.592\\
    & LexRank                 & ~29.85 &  5.87 & 17.56 & \light{11.84} &\U{.840}&.382&.525&.518\\
    & LexRank\subtext{SENTI}  & ~30.56 &  4.75 & 17.19 & \light{12.11} &.820&.441&.574&.572\\
    & LexRank\subtext{BERT}   & ~31.41 &  5.05 & 18.12 & \light{13.29} &.823&.380&.520&.500\\
    \midrule
                                                            
    \parbox[t]{1mm}{\multirow{3}{*}{\rotatebox[origin=c]{90}{\scriptsize{Abstract}}}}
    & Opinosis                & ~28.76 & 4.57  & 15.96 & \light{11.68} &.791&.446&.570&.561\\
    & MeanSum                 & ~34.95 & 7.49  & 19.92 & \light{14.52} &\textbf{.845}&.477&.610&.479\\
    & Copycat                 & ~\U{36.66} & \U{8.87}  & \U{20.90} & \light{14.15}&\U{.840}&.566&.676&\U{.446}\\
    \midrule                                                

    \multicolumn{2}{l@{~}|}{QT} & 
      ~\textbf{38.66} & \textbf{10.22} & \textbf{21.90} & \light{14.26} 
    & \U{.843}&\U{.689}&\textbf{.758}&\textbf{.430}\\
    \multicolumn{2}{@{}c@{~}|}{\textsl{~~~~w/o 2-step samp.}} & ~\U{37.82} & \U{9.13} & 20.10 & \light{13.88} 
    & \U{.833}&.680&\U{.748}&\U{.439}\\ \midrule
    \multicolumn{2}{@{}l@{}|}{Human Up. Bound} & ~49.80 & 18.80 & 29.19 & \light{34.58} 
    &.829&.862&.845&.264\\
    \bottomrule[1pt]
	\end{tabular}
    \caption{Summarization results on \corpus. Best system
      (shown in \textbf{boldface}) significantly outperforms all
      comparison systems, except where \U{underlined} ($p < 0.05$;
      paired bootstrap resampling;
      \citealp{koehn-2004-statistical}). We exclude Oracle systems from
      comparisons as they access gold summaries at test time.
      RL\subtext{ASP} is the
      Rouge-L of general summarizers against gold aspect
      summaries. AC and SC are shorthands for Aspect Coverage and
      Sentiment Consistency. Subscripts P and R refer to
      precision and recall, and F1 is their harmonic mean. MSE is mean
      squared error (lower is better).}
	\label{tab:space-general}
\end{minipage}
&
\begin{minipage}[b]{2.41in}
  \begin{tabular}{@{}l@{~}|@{~}c@{~~}c@{~~}c@{~~}|@{~}c@{~}|c@{}}
    \toprule[1pt]
    \yelp & R1    & R2   & RL & AC\ssubtext{F1} & SC\ssubtext{MSE}  \\
    \midrule[1pt]
    Random                  & 23.04 & 2.44 & 13.44             & .551 & .612 \\
    Centroid\ssubtext{BERT}  & 24.78 & 2.64 & 14.67             & .691 & .523 \\
    Oracle\ssubtext{BERT}    & \light{27.38} & \light{3.75} & \light{15.92} &
    \light{.703} & \light{.507}\\
    LexRank\ssubtext{BERT}   & 26.46 & 3.00 & 14.36             & .601 & .541 \\
    Opinosis                & 24.88 & 2.78 & 14.09             & .672 & .552 \\
    MeanSum                 & \U{28.46} & \U{3.66} & 15.57     &\U{.713}&\U{.510} \\
    Copycat                 & \textbf{29.47} & \textbf{5.26} & \textbf{18.09} & \textbf{.728} & \U{.495} \\
    \midrule
    QT                      & \U{28.40} & \U{3.97} & 15.27 & \U{.722} & \textbf{.490}\\
    \midrule[1pt]
    \multicolumn{4}{c}{\vspace*{-.11cm}}\\
    \midrule[1pt]
    \amazon & R1    & R2   & RL  & AC\ssubtext{F1} & SC\ssubtext{MSE}  \\
    \midrule[1pt]
    Random                  & 27.66 & 4.72 & 16.95 & .580 & .602 \\
    Centroid\ssubtext{BERT}  & 29.94 & 5.19 & 17.70 & .702 & .599 \\
    Oracle\ssubtext{BERT}    & \light{31.69} & \light{6.47} & \light{19.25} &
    \light{.725} & \light{.512} \\
    LexRank\ssubtext{BERT}   & \U{31.47} & 5.07 & 16.81 & .663 & .541 \\
    Opinosis                & 28.42 & 4.57 & 15.50 & .614 & .580 \\
    MeanSum                 & 29.20 & 4.70 & \U{18.15} & .710 & \U{.525}\\
    CopyCat                 & \U{31.97} & \U{5.81} & \textbf{20.16} & \U{.731} & \U{.510}\\
    \midrule
    QT                      & \textbf{34.04} & \textbf{7.03} & \U{18.08} & \textbf{.739} & \textbf{.508} \\
    \bottomrule[1pt]
	\end{tabular}
  \caption{Summarization results on \yelp{} and \amazon. Best system, shown
  in \textbf{boldface}, is significantly better than all comparison systems,
  except where \U{underlined} ($p < 0.05$; paired bootstrap resampling;
  \citealp{koehn-2004-statistical}).}
	\label{tab:yelp-amazon}
\end{minipage}
\end{tabular}
\end{table*}

\begin{table}[t]
	\footnotesize
	\centering
	\begin{tabular}{llllr}
    \toprule[1pt]
      & Inform.        & Coherent       & Concise        & Redund.        \\
    \midrule[1pt]
    Centroid & $+$36.0           & $-$57.3          & $-$60.7          & $-$12.7~~~\\
    LexRank  & $-$52.7           & $-$38.0          & $-$44.7          &  $-$1.3~~ \\
    MeanSum  & $-$23.3           & $+$26.7          & $+$28.7          &  $+$3.3~~ \\
    Copycat  & $-$10.7           & \textbf{$+$34.7} & $+$38.0          &  $-$3.3~~ \\
    QT       & \textbf{$+$50.7$^*$} & $+$34.0$^\dagger$
             & \textbf{$+$38.7}$^\dagger$ & \textbf{$+$18.0}$^*$ \\
    \bottomrule[1pt]
	\end{tabular}
  \caption{Best-Worst Scaling human study on \corpus{}. (*): significant
  difference to all models; ($\dagger$): significant difference to all models,
  except Copycat (one-way ANOVA with posthoc Tukey HSD test $p < 0.05$).}
	\label{tab:human-bws}
\end{table}

Table~\ref{tab:space-general} reports ROUGE scores on \corpus{} (test
set) for the general summarization task.  QT's popularity-based
extraction algorithm shows strong summarization capabilities
outperforming all comparison systems (differences in ROUGE are
statistically significant against all models but Copycat).  This
is a welcome result, considering that QT is an extractive method and
does not benefit from the compression and rewording capabilities of
abstractive summarizers. Moreover, as we discuss in
Section~\ref{sec:analysis}, QT is less data-hungry than other neural
models: it achieves the same level of performance even when trained on
5\% of the dataset. We also show in Table~\ref{tab:space-general}
(fourth block) that the proposed two-step sampling method yields better
extractive summaries compared to simply selecting the sentences nearest to the
most popular clusters.

Aspect coverage and sentiment consistency
results are also encouraging for QT which consistently scores highly on both
metrics, while baselines show mixed results.  We also compared (using
ROUGE-L) \emph{general} system summaries against reference
\emph{aspect} summaries. The results in Table~\ref{tab:space-general}
(column RL\subtext{ASP}) confirm that aspect summarization requires
tailor-made methods. Unsurprisingly, all systems are inferior
to the human upper bound (i.e.,~inter-annotator ROUGE and aspect-based metrics),
suggesting ample room for improvement.

QT's ability for general opinion summarization is further demonstrated
in Table~\ref{tab:yelp-amazon} which reports results on the \yelp{}
and \amazon{} datasets. We present the strongest baselines,
i.e.,~Centroid\subtext{BERT}, LexRank\subtext{BERT},
Oracle\subtext{BERT}, and the abstractive Opinosis, MeanSum, and
Copycat.  On \yelp, QT performs on par with MeanSum, but worse than
Copycat. However, it is important to note that, in contrast to
\corpus, \yelp's reference summaries were purposely written using
first-person narrative giving an advantage to review-like summaries of
abstractive methods. On \amazon, QT outperforms all methods on
\mbox{ROUGE-1/2}, but comes second to Copycat on ROUGE-L. This follows
a trend seen across all datasets, where abstractive systems appear
relatively stronger in terms of ROUGE-L compared to ROUGE-1/2. We partly
attribute this to their ability to fuse opinions into fluent
sentences, thus matching longer reference sequences.

Besides automatic evaluation, we conducted a user study to verify the
utility of the generated summaries.  We produced general summaries
from five systems (QT, Copycat, MeanSum, LexRank\subtext{BERT}
Centroid\subtext{BERT}) for all entities in \corpus's test set. For
every entity and pair of systems, we showed to three human judges a
gold-standard summary for reference, and the two system summaries. We
asked them to select the \emph{best} summary according to four
criteria: \emph{informativeness} (useful opinions, consistent with
reference), \emph{coherence} (easy to read, avoids contradictions),
\emph{conciseness} (useful in a few words), and
\emph{non-redundancy} (no repetitions). The systems' scores were
computed using \emph{Best-Worst Scaling} \citep{louviere2015best},
with values ranging from $-$100 (unanimously worst) to $+$100
(unanimously best). As shown in Table~\ref{tab:human-bws},
participants rate QT favorably over all baselines in terms of
informativeness, conciseness and lack of redundancy, with slight
preference for Copycat summaries with respect to coherence
(statistical significance information in caption). QT captures
essential opinions effectively, whereas there is room for improvement
in terms of summary cohesion.

\begin{table*}[t]
	\footnotesize
	\centering
    \begin{tabular}{@{}cl|cccccc|ccc|c@{}}
    \toprule[1pt]
      & & \multicolumn{6}{c|}{ROUGE-L} & R1 & R2 & RL & $\!\!\!$SC\ssubtext{MSE} \\ 
    \multicolumn{2}{c|}{\corpus{} \textsc{[Aspect]}} 
      & \textsl{Building} 
      & \textsl{Cleanliness} 
      & \textsl{Food} 
      & \textsl{Location} 
      & \textsl{Rooms} 
      & \textsl{Service} & 
    \multicolumn{3}{c|}{\textbf{Average}} &\\
    \midrule[1pt]

    \parbox[t]{1mm}{\multirow{3}{*}{\rotatebox[origin=c]{90}{\scriptsize{via BERT}}}}\hspace{-2mm}
    & MeanSum\ssubtext{ASP}
      &13.25&19.24&13.01&18.41&17.81&20.40&23.24&3.72&17.02&.235 \\
    & Copycat\ssubtext{ASP}
      &\textbf{17.10}&15.90&14.53&20.31&17.30&20.05&24.95&4.82&17.53&.274\\
    & LexRank\ssubtext{ASP}
      &\U{14.73}&\U{25.10}&\U{17.56}&\U{23.28}&18.24&\U{26.01}&\U{27.72} &
      \U{7.54} & \U{20.82}&\U{.206}\\
    \midrule
    & QT\ssubtext{ASP}
      &\U{16.45}&\textbf{25.12}&\textbf{17.79}&\textbf{23.63}&\textbf{21.61}&\textbf{26.07}&
      \textbf{28.95} & \textbf{8.34} & \textbf{21.77} & \textbf{.204}\\
    \midrule
      & Human & 40.33&38.76&33.63&35.23&29.25&30.31& 44.86 & 18.45 & 34.58 &.153 \\
    \bottomrule[1pt]
	\end{tabular}
  \vspace{-2mm}
	\caption{Aspect summarization results on \corpus. Best model shown in
	\textbf{boldface}. All differences to best model are statistically
	significant, except where \U{underlined}
	($p < 0.05$; paired bootstrap resampling; \citealp{koehn-2004-statistical}).}
	\label{tab:space-aspect}
\end{table*}

\begin{table}[t]
  \footnotesize
  \centering
  \begin{tabular}{lccc}
    \toprule[1pt]
    \multicolumn{4}{r}{\textbf{Does the summary discuss the specified aspect?}}\\
    \multicolumn{2}{r}{\hspace{20mm}Exclusively}   & Partially     & No              \\
    \midrule[1pt]
    QT\ssubtext{GEN}& ~~1.1              & 72.0               & 26.9            \\
    \midrule
    Copycat\ssubtext{ASP}        &  ~~6.7               & 45.3               & 48.0            \\
    MeanSum\ssubtext{ASP}        & 21.8               & 37.3               & 40.9            \\
    LexRank\ssubtext{ASP}        & 48.2               & 28.0               & 23.8            \\
    QT\ssubtext{ASP}             & \textbf{58.7}      & \textbf{32.7}      & \textbf{~~8.7}    \\
    \bottomrule[1pt]
  \end{tabular}
  \vspace{-1mm}
  \caption{User study on aspect-specific summaries. In the
  ``\textsl{Exclusively}'' column, QT's difference over all models is
  statistically significant ($p < 0.05$; $\chi^2$ test).}
  \label{tab:human-aspect}
\end{table}

\begin{table}[t]
  \footnotesize
  \centering
  \begin{tabular}{lp{.14\columnwidth}p{.14\columnwidth}p{.14\columnwidth}p{.07\columnwidth}}
    \toprule[1pt]
    \corpus & \multicolumn{4}{r}{\textbf{Proportion of \corpus's train data used~~}}\\
    \textsc{\footnotesize{[General]}} & 5\% & 10\% & 50\% & 100\% \\
    \midrule[1pt]
    Copycat & 26.1 & 26.2 & 31.8 & 36.7 \\
    QT & 36.9 & 37.1 & 37.7 & 38.7 \\
    \bottomrule[1pt]
  \end{tabular}
  \vspace{-2mm}
  \caption{ROUGE-1 on \corpus{} for varying train set sizes.}
  \label{tab:train-pct}
\end{table}

\subsection{Results: Aspect-specific Summarization}

There is no existing unsupervised system for aspect-specific opinion
summarization. Instead, we use the power of BERT 
\citep{devlin-etal-2019-bert} to enable aspect summarization for our
baselines. Specifically, we obtain BERT sentence vectors (average of token
vectors) for input sentences $X_e$, which we cluster via k-means. We then
replicate the cluster-to-aspects mapping used by QT, as described in
Equations~\mbox{(\ref{eq:asp-prob})--(\ref{eq:asp-codes})}: each cluster is
mapped to exactly one aspect, according to the probability of finding the
pre-defined aspect-denoting keywords in the sentences assigned to it. As a
result, we obtain non-overlapping and aspect-specific sets of input sentences
$\{X_e^{(a_1)}, X_e^{(a_2)}, \dots \}$. For aspect $a_i$, we create
{aspect-filtered} input reviews, by concatenating sentences in
$X_e^{(a_i)}$ based on the reviews they originated from. The filtered reviews of
each aspect are given as input to general summarizers (LexRank, MeanSum and
Copycat), thus producing aspect summaries. QT and all baselines use the same
aspect keywords, which we sourced from a held-out set of reviews, not
included in \corpus.

Table~\ref{tab:space-aspect} shows results on \corpus{}, for individual aspects,
and on average. QT outperforms baselines in all aspects, except
\textsl{building}, with significant improvements against Copycat and Meansum in
terms of ROUGE and sentiment consistency.  The abstractive methods struggle to
generate summaries restricted to the aspect in question.

To verify this, we ran a second judgement elicitation study.
We used summaries from competing aspect summarizers (QT\subtext{ASP},
Copycat\subtext{ASP}, MeanSum\subtext{ASP}, and LexRank\subtext{ASP})
for all six aspects, as well as QT's general summaries. A summary was
shown to three participants, who were asked whether it discussed the
specified aspect \emph{exclusively}, \emph{partially}, or \emph{not at
all}.  Table~\ref{tab:human-aspect} shows that 58.7\% of QT
aspect-specific summaries discuss the specified aspect exclusively,
while only 8.7\% of the summaries fail to mention the aspect.
LexRank\subtext{ASP} follows with 23.8\% of its summaries failing to
mention the aspect, while the abstractive models performed
significantly worse.

\begin{figure}[t]
  \centering
  \includegraphics[width=\columnwidth]{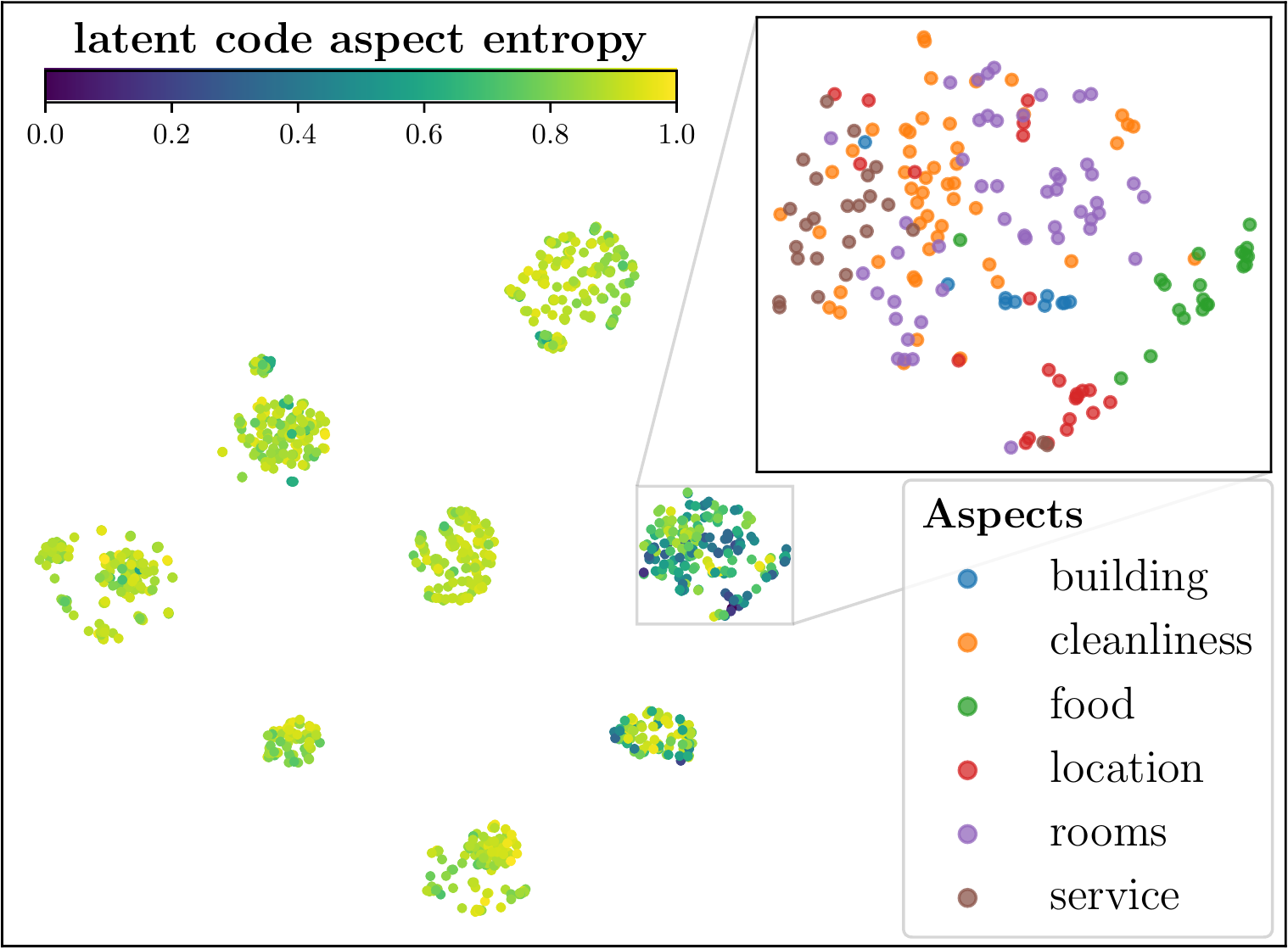}
  \caption{t-SNE projection of the quantized space of an eight-head QT
  trained on \corpus{}, showing all 1024 learned latent codes (best viewed in
  color). Darker codes correspond to lower \emph{aspect entropy}, our proposed
  measure of high aspect-specificity.
  Zooming in the aspect sub-space uncovers good aspect separation.}
  \label{fig:subspaces}
\end{figure}

\begin{table*}[t]
  \centering
  \footnotesize
  \begin{tabular}{p{.28\textwidth}|p{.22\textwidth}|p{.17\textwidth}|p{.23\textwidth}}
    \toprule[1pt]
    \textbf{Human} & \textbf{QT} & \textbf{MeanSum} & \textbf{Copycat} \\
    \midrule[1pt]
    All staff members were friendly, accommodating, and helpful. The hotel and
    room were very clean. The room had modern charm and was nicely remodeled.
    The beds are extremely comfortable. The rooms are quite with wonderful beach
    views. The food at Hash, the restaurant in lobby, was fabulous. The location
    is great, very close to the beach. It's a longish walk to Santa Monica. The
    price is very affordable. &

    Great hotel. We liked our room with an ocean view. The staff were friendly
    and helpful. There was no balcony. The location is perfect. Our room was
    very quiet. I would definitely stay here again. You're one block from the
    beach. So it must be good! Filthy hallways. Unvacuumed room. 
    Pricy, but well worth it. &

    It was a great stay! The food at the hotel is great for the price. I can't
    believe the noise from the street is very loud and the traffic is not so
    great, but that is not a problem. The restaurant was great and the food is
    excellent. &

    This hotel is in a great location, just off the beach. The staff was very
    friendly and helpful. We had a room with a view of the beach and ocean. The
    only problem was that our room was on the 4th floor with a view of the
    ocean. If you are looking for a nice place to sleep then this is the place
    for you. \\
    \bottomrule[1pt]
  \end{tabular}
  \caption{Four general opinion summaries for the same hotel: One human-written
  and three from competing models.}
  \vspace{2mm}
  \label{tab:outputs}
\end{table*}

\begin{figure}[h!]
  \centering
  \includegraphics[width=.99\columnwidth]{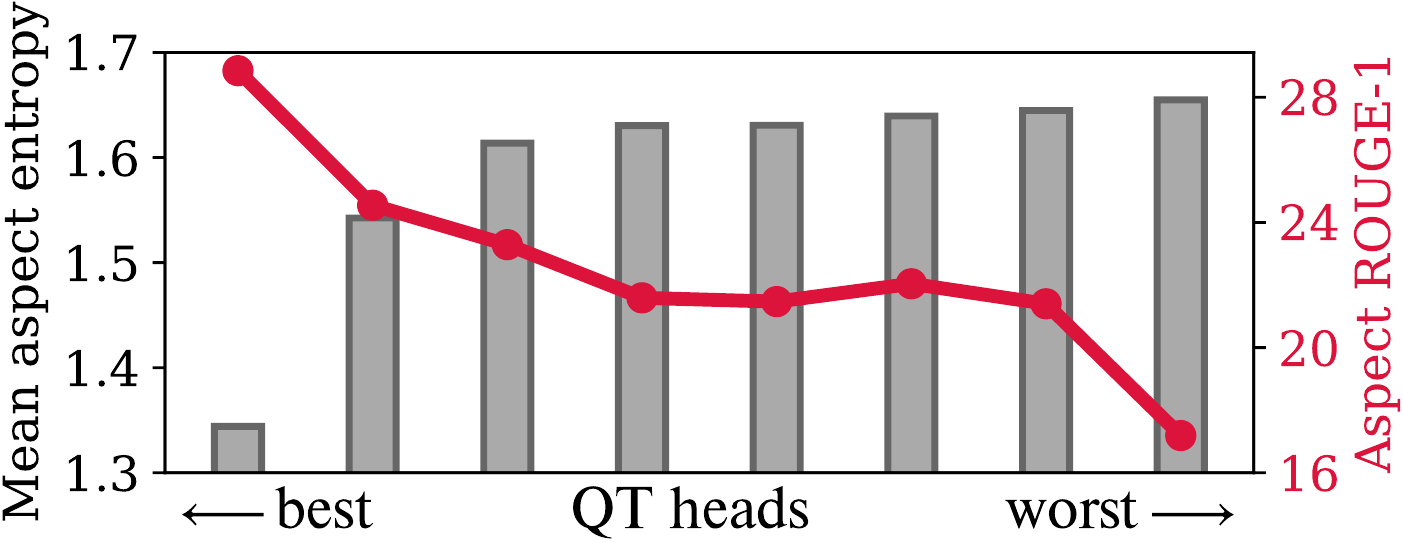}
  \caption{Mean aspect entropies (bars) for each of QT's head
  sub-spaces and corresponding aspect ROUGE-1 scores for the summaries produced by
  each head (line).}
  \label{fig:entropy}
\end{figure}

\subsection{Further Analysis}
\label{sec:analysis}

\paragraph{Training Efficiency}
Table~\ref{tab:train-pct} shows ROUGE-1 scores for QT and Copycat on
\corpus{}, when trained on different portions of the training set
(randomly downsampled and averaged over 5 runs). QT exhibits
impressive data efficiency; when trained on 5\% of data, it
performs comparably to a Copycat summarizer that has been trained on
the full corpus.

\begin{table*}[t]
  \centering
  \footnotesize
  \begin{tabular}{@{}p{16cm}@{}}
    \toprule[1pt]
    \textbf{Building:}
    Bright colors, skateboards, butterfly chairs and a grand ocean/boardwalk view
    (always entertaining). There is a small balcony, but there's only a small
    glass divider between your neighbor's balcony. \\

    \midrule

    \textbf{Food: } 
    We had a great breakfast at Hash too! The restaurant was amazing.
    Lots of good restaurants within walking distance and some even deliver. The
    roof bar was the icing on the cake.\\

    \midrule

    \textbf{Location:}
    The location is perfect. The hotel is very central. The hotel itself is
    in a great location. We hardly venture far as everything we need is within
    walking distance, but for the sightseers the buses are on the doorstep. \\
\midrule
    \textbf{Cleanliness:}
    Our room was very clean and comfortable. The room was clean and retrofitted
    with all the right amenities. Our room was very
    large, clean, and artfully decorated. \\
\midrule
    \textbf{Rooms:}
    The room was spacious and had really cool furnishings, and the beds were
    comfortable. The room's were good, and we had a free upgrade for one of them
    (for a Facebook 'like!) A+ for the bed and pillows. \\

    \midrule
    \textbf{Service:}
    The staff is great. The staff were friendly and helpful. The hotel staff
    were friendly and provided us with great service. Each member of the staff
    was friendly and attentive. The staff excel and nothing is ever too much
    trouble. \\
    \bottomrule[1pt]
  \end{tabular}
  \caption{Aspect summaries extracted by QT.}
  \label{tab:aspect-outputs}
\end{table*}

\paragraph{Visualizing Sub-spaces}
We present a visual demonstration of QT's quantized sub-spaces in
Figure~\ref{fig:subspaces}. We used t-SNE \citep{tsne} to project the latent
code vectors onto two dimensions. The latent codes produced by QT's
eight heads are clearly grouped in eight separate sub-spaces. The aspect
sub-space (shown in square) was detected automatically, as it displayed
the lowest mean aspect entropy (darker color). Zooming into its latent codes
uncovers reasonable aspect separation, an impressive result
considering that the model received no aspect-specific supervision.

\paragraph{Mean Aspect Entropy}
Figure~\ref{fig:entropy} further illustrates the effectiveness of aspect entropy
for detecting the head sub-space that best separates aspect-specific sentences.
Each gray bar shows the mean aspect entropy for the codes produced by one of
QT's eight heads. One of the heads (leftmost) exhibits much lower entropy, indicating a
strong confidence for aspect membership within its latent codes. We confirm this
enables better aspect summarization by generating aspect summaries using each
head, and plotting the obtained ROUGE-1 scores.

\paragraph{System Output}

Finally, we show gold-standard and system-generated general summaries
in Table~\ref{tab:outputs}, as well as QT aspect summaries in
Table~\ref{tab:aspect-outputs}.

\section{Conclusions}
\label{sec:conc}

We presented a novel opinion summarization system based on the Quantized
Transformer that requires no reference summaries for training, and is able to
extract general and aspect summaries from large groups of input reviews. QT is
trained through sentence reconstruction and learns a rich encoding space, paired
with a clustering component based on vector quantized variational autoencoders.
At summarization time, we exploit the characteristics of the quantized space, to
identify those clusters that correspond to the input's most popular opinions,
and extract the sentences that best represent them. Moreover, we used the
multi-head representations of the model, and no further training, to detect the
encoding sub-space that best separates aspects, enabling aspect-specific
summarization. We also collected \corpus, a new opinion summarization corpus
which we hope will inform and inspire further research. 

Experimental results on \corpus{} and popular benchmarks reveal that
our system is able to produce informative summaries which cover all or
individual aspects of an entity. In the future, we would like to
utilize the QT framework in order to generate abstractive
summaries. We could also exploit QT's multi-head semantics more
directly, and further improve it through weak supervision or
multi-task objectives. Finally, although we focused on opinion
summarization, it would be interesting to see if the proposed model
can be applied to other multi-document summarization tasks.

\section*{Acknowledgements} 
We thank the anonymous reviewers for their feedback and the action editor, Jing
Jiang, for her comments.  We gratefully  acknowledge the  support  of  the
European  Research   Council (Lapata; award  number  681760,  ``Translating
Multiple Modalities into Text''). We also thank Wang-Chiew Tan for her valuable
input.

\bibliographystyle{acl_natbib}
\bibliography{tacl2020}

\begin{thebibliography}{45}
\expandafter\ifx\csname natexlab\endcsname\relax\def\natexlab#1{#1}\fi

\bibitem[{Amplayo and Lapata(2019)}]{amplayo-lapata-2019-informative}
Reinald~Kim Amplayo and Mirella Lapata. 2019.
\newblock \href {https://arxiv.org/abs/1909.02322} {Informative and
  controllable opinion summarization}.
\newblock \emph{arXiv preprint arXiv:1909.02322v1}.

\bibitem[{Amplayo and Lapata(2020)}]{amplayo-lapata-2020-unsupervised}
Reinald~Kim Amplayo and Mirella Lapata. 2020.
\newblock \href {https://doi.org/10.18653/v1/2020.acl-main.175} {Unsupervised
  opinion summarization with noising and denoising}.
\newblock In \emph{Proceedings of the 58th Annual Meeting of the Association
  for Computational Linguistics}, pages 1934--1945, Online. Association for
  Computational Linguistics.

\bibitem[{Angelidis and Lapata(2018)}]{angelidis-lapata-2018-summarizing}
Stefanos Angelidis and Mirella Lapata. 2018.
\newblock \href {https://doi.org/10.18653/v1/D18-1403} {Summarizing opinions:
  Aspect extraction meets sentiment prediction and they are both weakly
  supervised}.
\newblock In \emph{Proceedings of the 2018 Conference on Empirical Methods in
  Natural Language Processing}, pages 3675--3686, Brussels, Belgium.
  Association for Computational Linguistics.

\bibitem[{Bengio et~al.(2013)Bengio, L{\'e}onard, and
  Courville}]{bengio-etal-2013-estimating}
Yoshua Bengio, Nicholas L{\'e}onard, and Aaron Courville. 2013.
\newblock \href {https://arxiv.org/abs/1308.3432} {Estimating or propagating
  gradients through stochastic neurons for conditional computation}.
\newblock \emph{arXiv preprint arXiv:1308.3432v1}.

\bibitem[{Bra{\v{z}}inskas et~al.(2020)Bra{\v{z}}inskas, Lapata, and
  Titov}]{bravzinskas-etal-2020-unsupervised}
Arthur Bra{\v{z}}inskas, Mirella Lapata, and Ivan Titov. 2020.
\newblock \href {https://doi.org/10.18653/v1/2020.acl-main.461} {Unsupervised
  opinion summarization as copycat-review generation}.
\newblock In \emph{Proceedings of the 58th Annual Meeting of the Association
  for Computational Linguistics}, pages 5151--5169, Online. Association for
  Computational Linguistics.

\bibitem[{Cao et~al.(2015)Cao, Wei, Dong, Li, and Zhou}]{cao2015ranking}
Ziqiang Cao, Furu Wei, Li~Dong, Sujian Li, and Ming Zhou. 2015.
\newblock \href
  {https://www.aaai.org/ocs/index.php/AAAI/AAAI15/paper/view/9414} {Ranking
  with recursive neural networks and its application to multi-document
  summarization}.
\newblock In \emph{Proceedings of the Twenty-Ninth AAAI Conference on
  Artificial Intelligence}, AAAI'15, pages 2153--2159. AAAI Press.

\bibitem[{Carenini et~al.(2006)Carenini, Ng, and Pauls}]{carenini2006multi}
Giuseppe Carenini, Raymond Ng, and Adam Pauls. 2006.
\newblock \href {https://www.aclweb.org/anthology/E06-1039} {Multi-document
  summarization of evaluative text}.
\newblock In \emph{11th Conference of the {E}uropean Chapter of the Association
  for Computational Linguistics}, Trento, Italy. Association for Computational
  Linguistics.

\bibitem[{Cheng and Lapata(2016)}]{cheng-lapata-2016-neural}
Jianpeng Cheng and Mirella Lapata. 2016.
\newblock \href {https://doi.org/10.18653/v1/P16-1046} {Neural summarization by
  extracting sentences and words}.
\newblock In \emph{Proceedings of the 54th Annual Meeting of the Association
  for Computational Linguistics (Volume 1: Long Papers)}, pages 484--494,
  Berlin, Germany. Association for Computational Linguistics.

\bibitem[{Chu and Liu(2019)}]{chu-liu-2019-meansum}
Eric Chu and Peter Liu. 2019.
\newblock \href {http://proceedings.mlr.press/v97/chu19b.html} {{M}ean{S}um: A
  neural model for unsupervised multi-document abstractive summarization}.
\newblock In \emph{Proceedings of the 36th International Conference on Machine
  Learning}, volume~97 of \emph{Proceedings of Machine Learning Research},
  pages 1223--1232, Long Beach, California, USA. PMLR.

\bibitem[{Coavoux et~al.(2019)Coavoux, Elsahar, and
  Gall{\'e}}]{coavoux-etal-2019-unsupervised}
Maximin Coavoux, Hady Elsahar, and Matthias Gall{\'e}. 2019.
\newblock \href {https://doi.org/10.18653/v1/D19-5405} {Unsupervised
  aspect-based multi-document abstractive summarization}.
\newblock In \emph{Proceedings of the 2nd Workshop on New Frontiers in
  Summarization}, pages 42--47, Hong Kong, China. Association for Computational
  Linguistics.

\bibitem[{Devlin et~al.(2019)Devlin, Chang, Lee, and
  Toutanova}]{devlin-etal-2019-bert}
Jacob Devlin, Ming-Wei Chang, Kenton Lee, and Kristina Toutanova. 2019.
\newblock \href {https://doi.org/10.18653/v1/N19-1423} {{BERT}: Pre-training of
  deep bidirectional transformers for language understanding}.
\newblock In \emph{Proceedings of the 2019 Conference of the North {A}merican
  Chapter of the Association for Computational Linguistics: Human Language
  Technologies, Volume 1 (Long and Short Papers)}, pages 4171--4186,
  Minneapolis, Minnesota. Association for Computational Linguistics.

\bibitem[{Di~Fabbrizio et~al.(2014)Di~Fabbrizio, Stent, and
  Gaizauskas}]{difabbrizio2014hybrid}
Giuseppe Di~Fabbrizio, Amanda Stent, and Robert Gaizauskas. 2014.
\newblock \href {https://doi.org/10.3115/v1/W14-4408} {A hybrid approach to
  multi-document summarization of opinions in reviews}.
\newblock In \emph{Proceedings of the 8th International Natural Language
  Generation Conference ({INLG})}, pages 54--63, Philadelphia, Pennsylvania.
  Association for Computational Linguistics.

\bibitem[{Erkan and Radev(2004)}]{erkan-radev-2004-lexrank}
G\"{u}nes Erkan and Dragomir~R. Radev. 2004.
\newblock \href {https://www.aaai.org/Papers/JAIR/Vol22/JAIR-2214.pdf}
  {Lexrank: Graph-based lexical centrality as salience in text summarization}.
\newblock \emph{J. Artif. Int. Res.}, 22(1):457--479.

\bibitem[{Ganesan et~al.(2010)Ganesan, Zhai, and
  Han}]{ganesan-etal-2010-opinosis}
Kavita Ganesan, ChengXiang Zhai, and Jiawei Han. 2010.
\newblock \href {https://www.aclweb.org/anthology/C10-1039} {{O}pinosis: A
  graph based approach to abstractive summarization of highly redundant
  opinions}.
\newblock In \emph{Proceedings of the 23rd International Conference on
  Computational Linguistics (Coling 2010)}, pages 340--348, Beijing, China.
  Coling 2010 Organizing Committee.

\bibitem[{Holtzman et~al.(2020)Holtzman, Buys, Forbes, and
  Choi}]{holtzman-etal-2019-curious}
Ari Holtzman, Jan Buys, Maxwell Forbes, and Yejin Choi. 2020.
\newblock \href {https://openreview.net/forum?id=rygGQyrFvH} {The curious case
  of neural text degeneration}.
\newblock In \emph{8th International Conference on Learning Representations,
  {ICLR} 2020, Virtual Conference, Apr 26th - May 1st}. OpenReview.net.

\bibitem[{Hu and Liu(2004)}]{hu-liu-2004-mining}
Minqing Hu and Bing Liu. 2004.
\newblock \href {https://doi.org/10.1145/1014052.1014073} {Mining and
  summarizing customer reviews}.
\newblock In \emph{Proceedings of the Tenth ACM SIGKDD International Conference
  on Knowledge Discovery and Data Mining}, KDD '04, pages 168--177, New York,
  NY, USA. Association for Computing Machinery.

\bibitem[{Huy~Tien et~al.(2019)Huy~Tien, Tung~Thanh, and
  Minh~Le}]{huy-tien-etal-2019-opinions}
Nguyen Huy~Tien, Le~Tung~Thanh, and Nguyen Minh~Le. 2019.
\newblock \href {https://doi.org/10.26615/978-954-452-056-4_058} {Opinions
  summarization: Aspect similarity recognition relaxes the constraint of
  predefined aspects}.
\newblock In \emph{Proceedings of the International Conference on Recent
  Advances in Natural Language Processing (RANLP 2019)}, pages 487--496, Varna,
  Bulgaria. INCOMA Ltd.

\bibitem[{Isonuma et~al.(2019)Isonuma, Mori, and
  Sakata}]{isonuma-etal-2019-unsupervised}
Masaru Isonuma, Junichiro Mori, and Ichiro Sakata. 2019.
\newblock \href {https://doi.org/10.18653/v1/P19-1206} {Unsupervised neural
  single-document summarization of reviews via learning latent discourse
  structure and its ranking}.
\newblock In \emph{Proceedings of the 57th Annual Meeting of the Association
  for Computational Linguistics}, pages 2142--2152, Florence, Italy.
  Association for Computational Linguistics.

\bibitem[{Kaiser and Bengio(2018)}]{Kaiser2018DiscreteAF}
Lukasz Kaiser and Samy Bengio. 2018.
\newblock \href {https://arxiv.org/abs/1801.09797} {Discrete autoencoders for
  sequence models}.
\newblock \emph{ArXiv}, abs/1801.09797v1.

\bibitem[{Kingma and Welling(2014)}]{Kingma2014AutoEncodingVB}
Diederik~P. Kingma and Max Welling. 2014.
\newblock \href {https://arxiv.org/abs/1312.6114} {Auto-encoding variational
  bayes}.
\newblock \emph{CoRR}, abs/1312.6114v10.

\bibitem[{Koehn(2004)}]{koehn-2004-statistical}
Philipp Koehn. 2004.
\newblock \href {https://www.aclweb.org/anthology/W04-3250} {Statistical
  significance tests for machine translation evaluation}.
\newblock In \emph{Proceedings of the 2004 Conference on Empirical Methods in
  Natural Language Processing}, pages 388--395, Barcelona, Spain. Association
  for Computational Linguistics.

\bibitem[{Liu et~al.(2018)Liu, Saleh, Pot, Goodrich, Sepassi, Kaiser, and
  Shazeer}]{liu2018generating}
Peter~J. Liu, Mohammad Saleh, Etienne Pot, Ben Goodrich, Ryan Sepassi, Lukasz
  Kaiser, and Noam Shazeer. 2018.
\newblock \href {https://openreview.net/forum?id=Hyg0vbWC-} {Generating
  wikipedia by summarizing long sequences}.
\newblock In \emph{6th International Conference on Learning Representations,
  {ICLR} 2018, Vancouver, BC, Canada, April 30 - May 3, 2018, Conference Track
  Proceedings}. OpenReview.net.

\bibitem[{Louviere et~al.(2015)Louviere, Flynn, and Marley}]{louviere2015best}
Jordan~J Louviere, Terry~N Flynn, and Anthony Alfred~John Marley. 2015.
\newblock \href {https://doi.org/https://doi.org/10.1017/CBO9781107337855}
  {\emph{Best-worst scaling: Theory, methods and applications}}.
\newblock Cambridge University Press.

\bibitem[{Maddison et~al.(2017)Maddison, Mnih, and Teh}]{Maddison2017TheCD}
Chris~J. Maddison, Andriy Mnih, and Yee~Whye Teh. 2017.
\newblock \href {https://openreview.net/forum?id=S1jE5L5gl} {The concrete
  distribution: A continuous relaxation of discrete random variables}.
\newblock In \emph{International Conference on Learning Representations}.

\bibitem[{Marcheggiani et~al.(2014)Marcheggiani, T{\"a}ckstr{\"o}m, Esuli, and
  Sebastiani}]{marcheggiani2014hier}
Diego Marcheggiani, Oscar T{\"a}ckstr{\"o}m, Andrea Esuli, and Fabrizio
  Sebastiani. 2014.
\newblock \href
  {https://link.springer.com/chapter/10.1007/978-3-319-06028-6_23}
  {Hierarchical multi-label conditional random fields for aspect-oriented
  opinion mining}.
\newblock In \emph{Advances in Information Retrieval}, pages 273--285, Cham.
  Springer International Publishing.

\bibitem[{Miao et~al.(2020)Miao, Li, Wang, and Tan}]{miao2020snippext}
Zhengjie Miao, Yuliang Li, Xiaolan Wang, and Wang-Chiew Tan. 2020.
\newblock \href {https://doi.org/10.1145/3366423.3380144} {Snippext:
  Semi-supervised opinion mining with augmented data}.
\newblock In \emph{Proceedings of The Web Conference 2020}, WWW '20, pages
  617--628, New York, NY, USA. Association for Computing Machinery.

\bibitem[{Narayan et~al.(2018)Narayan, Cohen, and
  Lapata}]{narayan-etal-2018-ranking}
Shashi Narayan, Shay~B. Cohen, and Mirella Lapata. 2018.
\newblock \href {https://doi.org/10.18653/v1/N18-1158} {Ranking sentences for
  extractive summarization with reinforcement learning}.
\newblock In \emph{Proceedings of the 2018 Conference of the North {A}merican
  Chapter of the Association for Computational Linguistics: Human Language
  Technologies, Volume 1 (Long Papers)}, pages 1747--1759, New Orleans,
  Louisiana. Association for Computational Linguistics.

\bibitem[{van~den Oord et~al.(2017)van~den Oord, Vinyals, and
  kavukcuoglu}]{oord-etal-2017-neural}
Aaron van~den Oord, Oriol Vinyals, and koray kavukcuoglu. 2017.
\newblock \href
  {http://papers.nips.cc/paper/7210-neural-discrete-representation-learning.pdf}
  {Neural discrete representation learning}.
\newblock In \emph{Advances in Neural Information Processing Systems 30}, pages
  6306--6315. Curran Associates, Inc.

\bibitem[{Pan et~al.(2020)Pan, Yang, Zhou, Wang, Cai, and Liu}]{pan2020large}
Haojie Pan, Rongqin Yang, Xin Zhou, Rui Wang, Deng Cai, and Xiaozhong Liu.
  2020.
\newblock \href {https://doi.org/10.1145/3397271.3401439} {Large scale
  abstractive multi-review summarization (lsars) via aspect alignment}.
\newblock In \emph{Proceedings of the 43rd International ACM SIGIR Conference
  on Research and Development in Information Retrieval}, SIGIR '20, pages
  2337--2346, New York, NY, USA. Association for Computing Machinery.

\bibitem[{Pang and Lee(2008)}]{pang-lee-2008-opinion}
Bo~Pang and Lillian Lee. 2008.
\newblock \href {https://doi.org/10.1561/1500000011} {Opinion mining and
  sentiment analysis}.
\newblock \emph{Foundations and Trends® in Information Retrieval},
  2(1--2):1--135.

\bibitem[{Perez-Beltrachini et~al.(2019)Perez-Beltrachini, Liu, and
  Lapata}]{perez2019generating}
Laura Perez-Beltrachini, Yang Liu, and Mirella Lapata. 2019.
\newblock \href {https://doi.org/10.18653/v1/P19-1504} {Generating summaries
  with topic templates and structured convolutional decoders}.
\newblock In \emph{Proceedings of the 57th Annual Meeting of the Association
  for Computational Linguistics}, pages 5107--5116, Florence, Italy.
  Association for Computational Linguistics.

\bibitem[{Peyrard(2019)}]{peyrard-2019-simple}
Maxime Peyrard. 2019.
\newblock \href {https://doi.org/10.18653/v1/P19-1101} {A simple theoretical
  model of importance for summarization}.
\newblock In \emph{Proceedings of the 57th Annual Meeting of the Association
  for Computational Linguistics}, pages 1059--1073, Florence, Italy.
  Association for Computational Linguistics.

\bibitem[{Radev et~al.(2003)Radev, Teufel, Saggion, Lam, Blitzer, Qi,
  {\c{C}}elebi, Liu, and Drabek}]{radev-etal-2003-evaluation}
Dragomir~R. Radev, Simone Teufel, Horacio Saggion, Wai Lam, John Blitzer, Hong
  Qi, Arda {\c{C}}elebi, Danyu Liu, and Elliott Drabek. 2003.
\newblock \href {https://doi.org/10.3115/1075096.1075144} {Evaluation
  challenges in large-scale document summarization}.
\newblock In \emph{Proceedings of the 41st Annual Meeting of the Association
  for Computational Linguistics}, pages 375--382, Sapporo, Japan. Association
  for Computational Linguistics.

\bibitem[{Radford et~al.(2017)Radford, Jozefowicz, and
  Sutskever}]{radford-etal-2017-learning}
Alec Radford, Rafal Jozefowicz, and Ilya Sutskever. 2017.
\newblock \href {https://arxiv.org/abs/1704.01444} {Learning to generate
  reviews and discovering sentiment}.
\newblock \emph{arXiv preprint arXiv:1704.01444v2}.

\bibitem[{Rohrbach et~al.(2018)Rohrbach, Hendricks, Burns, Darrell, and
  Saenko}]{rohrbach-etal-2018-object}
Anna Rohrbach, Lisa~Anne Hendricks, Kaylee Burns, Trevor Darrell, and Kate
  Saenko. 2018.
\newblock \href {https://doi.org/10.18653/v1/D18-1437} {Object hallucination in
  image captioning}.
\newblock In \emph{Proceedings of the 2018 Conference on Empirical Methods in
  Natural Language Processing}, pages 4035--4045, Brussels, Belgium.
  Association for Computational Linguistics.

\bibitem[{Roy and Grangier(2019)}]{roy-grangier-2019-unsupervised}
Aurko Roy and David Grangier. 2019.
\newblock \href {https://doi.org/10.18653/v1/P19-1605} {Unsupervised
  paraphrasing without translation}.
\newblock In \emph{Proceedings of the 57th Annual Meeting of the Association
  for Computational Linguistics}, pages 6033--6039, Florence, Italy.
  Association for Computational Linguistics.

\bibitem[{Roy et~al.(2018)Roy, Vaswani, Neelakantan, and
  Parmar}]{roy-etal-2018-theory}
Aurko Roy, Ashish Vaswani, Arvind Neelakantan, and Niki Parmar. 2018.
\newblock \href {https://arxiv.org/abs/1805.11063} {Theory and experiments on
  vector quantized autoencoders}.
\newblock \emph{arXiv preprint arXiv:1805.11063v2}.

\bibitem[{See et~al.(2017)See, Liu, and Manning}]{see-etal-2017-get}
Abigail See, Peter~J. Liu, and Christopher~D. Manning. 2017.
\newblock \href {https://doi.org/10.18653/v1/P17-1099} {Get to the point:
  Summarization with pointer-generator networks}.
\newblock In \emph{Proceedings of the 55th Annual Meeting of the Association
  for Computational Linguistics (Volume 1: Long Papers)}, pages 1073--1083,
  Vancouver, Canada. Association for Computational Linguistics.

\bibitem[{Suhara et~al.(2020)Suhara, Wang, Angelidis, and
  Tan}]{suhara-etal-2020-opiniondigest}
Yoshihiko Suhara, Xiaolan Wang, Stefanos Angelidis, and Wang-Chiew Tan. 2020.
\newblock \href {https://doi.org/10.18653/v1/2020.acl-main.513}
  {{O}pinion{D}igest: A simple framework for opinion summarization}.
\newblock In \emph{Proceedings of the 58th Annual Meeting of the Association
  for Computational Linguistics}, pages 5789--5798, Online. Association for
  Computational Linguistics.

\bibitem[{Tian et~al.(2019)Tian, Yu, and Jiang}]{tian-etal-2019-aspect}
Yufei Tian, Jianfei Yu, and Jing Jiang. 2019.
\newblock \href {https://doi.org/10.1145/3357384.3358142} {Aspect and opinion
  aware abstractive review summarization with reinforced hard typed decoder}.
\newblock In \emph{Proceedings of the 28th ACM International Conference on
  Information and Knowledge Management}, CIKM '19, pages 2061--2064, New York,
  NY, USA. Association for Computing Machinery.

\bibitem[{{van der Maaten} and Hinton(2008)}]{tsne}
L.J.P. {van der Maaten} and G.E. Hinton. 2008.
\newblock \href
  {https://www.jmlr.org/papers/volume9/vandermaaten08a/vandermaaten08a.pdf}
  {Visualizing data using t-{SNE}}.
\newblock \emph{Journal of Machine Learning Research}, 9(nov):2579--2605.
\newblock Pagination: 27.

\bibitem[{Vaswani et~al.(2017)Vaswani, Shazeer, Parmar, Uszkoreit, Jones,
  Gomez, Kaiser, and Polosukhin}]{vaswani-etal-2017-attention}
Ashish Vaswani, Noam Shazeer, Niki Parmar, Jakob Uszkoreit, Llion Jones,
  Aidan~N Gomez, \L~ukasz Kaiser, and Illia Polosukhin. 2017.
\newblock \href
  {http://papers.nips.cc/paper/7181-attention-is-all-you-need.pdf} {Attention
  is all you need}.
\newblock In I.~Guyon, U.~V. Luxburg, S.~Bengio, H.~Wallach, R.~Fergus,
  S.~Vishwanathan, and R.~Garnett, editors, \emph{Advances in Neural
  Information Processing Systems 30}, pages 5998--6008. Curran Associates, Inc.

\bibitem[{Wang et~al.(2010)Wang, Lu, and Zhai}]{wang-etal-2010-latent}
Hongning Wang, Yue Lu, and Chengxiang Zhai. 2010.
\newblock \href {https://doi.org/10.1145/1835804.1835903} {Latent aspect rating
  analysis on review text data: A rating regression approach}.
\newblock In \emph{Proceedings of the 16th ACM SIGKDD International Conference
  on Knowledge Discovery and Data Mining}, KDD '10, pages 783--792, New York,
  NY, USA. Association for Computing Machinery.

\bibitem[{Wang et~al.(2020)Wang, Suhara, Nuno, Li, Li, Carmeli, Angelidis,
  Kandogann, and Tan}]{wang2020extreme}
Xiaolan Wang, Yoshihiko Suhara, Natalie Nuno, Yuliang Li, Jinfeng Li, Nofar
  Carmeli, Stefanos Angelidis, Eser Kandogann, and Wang-Chiew Tan. 2020.
\newblock \href {https://doi.org/10.1145/3366424.3383535} {{E}xtreme{R}eader:
  {A}n interactive explorer for customizable and explainable review
  summarization}.
\newblock In \emph{Companion Proceedings of the Web Conference 2020}, WWW '20,
  pages 176--180, New York, NY, USA. Association for Computing Machinery.

\bibitem[{Yasunaga et~al.(2017)Yasunaga, Zhang, Meelu, Pareek, Srinivasan, and
  Radev}]{yasunaga-etal-2017-graph}
Michihiro Yasunaga, Rui Zhang, Kshitijh Meelu, Ayush Pareek, Krishnan
  Srinivasan, and Dragomir Radev. 2017.
\newblock \href {https://doi.org/10.18653/v1/K17-1045} {Graph-based neural
  multi-document summarization}.
\newblock In \emph{Proceedings of the 21st Conference on Computational Natural
  Language Learning ({C}o{NLL} 2017)}, pages 452--462, Vancouver, Canada.
  Association for Computational Linguistics.

\end{thebibliography}

\end{document}